\documentclass{article}
\usepackage{pmode}

\usepackage[utf8]{inputenc} 
\usepackage[T1]{fontenc}    
\usepackage{hyperref}       
\usepackage{url}            
\usepackage{booktabs}       
\usepackage{amsfonts}       
\usepackage{nicefrac}       
\usepackage{microtype}      
\usepackage{xcolor}
\usepackage{lipsum}
\usepackage{fancyhdr}       
\usepackage{graphicx}       
\graphicspath{ {images/} } 
\usepackage{amsmath}
\usepackage{float}
\usepackage[font=footnotesize,labelfont=bf]{caption}
\usepackage{tabularx}
\usepackage{amsmath}
\usepackage{tabulary}
\usepackage{multirow}
\usepackage{caption}
\usepackage{subcaption}

\title{PMODE - Prototypical Mask based Object Dimension Estimation}

\author{
  Thariq Khalid \\
  Elm Company \\
  \texttt{tkadavil@elm.sa} \\
  \And
  Mohammed Yahya Hakami \\
  Elm Company \\
  \texttt{mohhakami@elm.sa} \\
   \And
  Riad Souissi \\
  Elm Company \\
  \texttt{rsouissi@elm.sa} \\
}

\def\plainkeywords{: Street View, Scene Understanding, Signboard, Autonomous Driving, Digital Mapping, Smart cities}

\hypersetup{%
  pdfkeywords={\plainkeywords},
}
\begin{document}

\maketitle

\begin{abstract}
Can a neural network estimate an object’s dimension in the wild? In this paper, we propose a method and deep learning architecture to estimate the dimensions of a quadrilateral object of interest in videos using a monocular camera. The proposed technique does not use camera calibration or handcrafted geometric features; however, features are learned with the help of coefficients of a segmentation neural network during the training process. A real-time instance segmentation-based Deep Neural Network with a ResNet50 backbone is employed, giving the object’s prototype mask and thus provides a region of interest to regress its dimensions. The instance segmentation network is trained to look at only the nearest object of interest. The regression is performed using an MLP head which looks only at the mask coefficients of the bounding box detector head and the prototype segmentation mask. We trained the system with three different random cameras achieving 22\% MAPE for the test dataset for the dimension estimation.
\end{abstract}
\section{Introduction}

Measuring the dimension of common objects of interest like buildings, footpaths, parking spaces, and shop signages is essential for municipalities worldwide for asset tracking and maintenance. Traditionally, humans used to obtain the dimensions of objects by measurements using tapes and necessary tools. But in the long run, using human labor to estimate these objects’ dimensions is very difficult. Very tall objects require ladders or similar helpful instruments to reach the object.  These procedures are risky, and they cause damage to the workers involved.  Today, there are laser-equipped devices in the market for objects that are out of reach from the human hand. However, this is a very tedious and time-consuming procedure. Usually, it needs multiple persons take the measurements. The expected error in this process is very high due to the data collection process, supervision, and review, as well as the error of the instruments used.

Computer Vision using Convolutional Neural Networks\cite{lecun1995convolutional} and deep learning can perform classification, detection, and segmentation of objects with high accuracy using state-of-the-art techniques. Deep learning has advanced to the level of self-driving cars fueled by data collected from the roads and eventually operated by many Neural Networks running in parallel and sequentially. Self-driving vehicles use the fusion of multiple sensor inputs from cameras, radars, and LiDARs. Long-range LiDARs and short-range LiDARs have become imperative to get the distance and depth of objects accurately. To navigate an environment, self-driving cars do not necessarily need to know the dimensions of objects they come across. They only need to detect the boundaries and navigate without collision. However, to make decisions about overtaking, gauging the distance to a target, generating occupancy grids, etc., they need to estimate the dimensions of particular objects. Such critical decision-making solutions should be real-time and quick so that the systems benefit the most from them.

Shop signages are common objects which can be found publicly on the roadsides. Measuring the dimensions of objects such as shop signage could  provide value in planning for smart city objectives related to real estate and compliance. In this work, we propose a Deep Neural Network architecture to estimate the dimensions of shop signages in videos. In the process, we also created a novel video dataset containing shop signages and their corresponding dimensions. The videos were collected from cities in two countries using cameras with varied settings from inside a moving car. During the model evaluation, the estimations are done using the video stream of a camera placed inside the moving car.

Following the architecture of YOLACT\cite{bolya2019yolact}, which is highly suitable for real-time Instance segmentation performance on videos, our deep neural network has a Resnet-50-based backbone\cite{he2016deep}. The backbone is connected to a feature pyramid. Our network then uses a prototype mask and the bounding box detector-based mask coefficients to predict the dimensions of the shop signage with the help of a Multi-Layer Perceptron. We achieve 22\% MAPE on the test dataset for the regression of dimensions. We also shed some light on the novel dataset we collected through our industrial process and provided the statistics.

\begin{figure*}
\begin{center}
\includegraphics[width=0.49\linewidth]{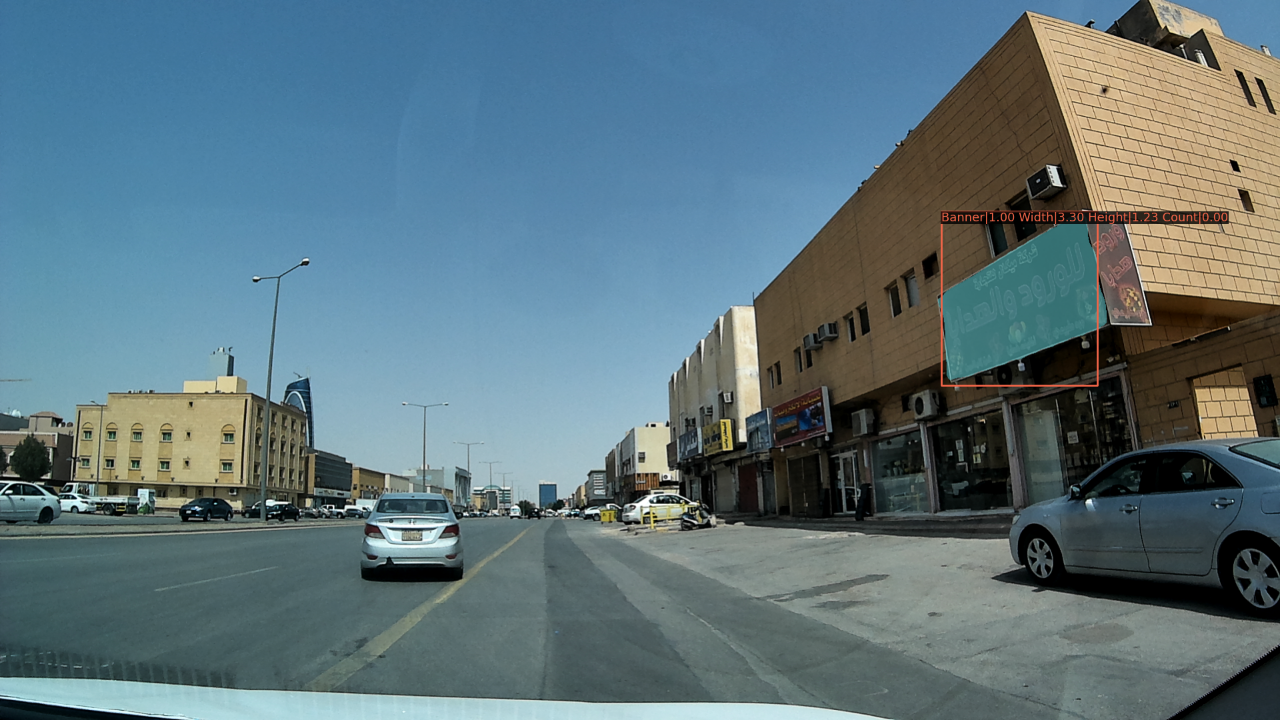}
\includegraphics[width=0.49\linewidth]{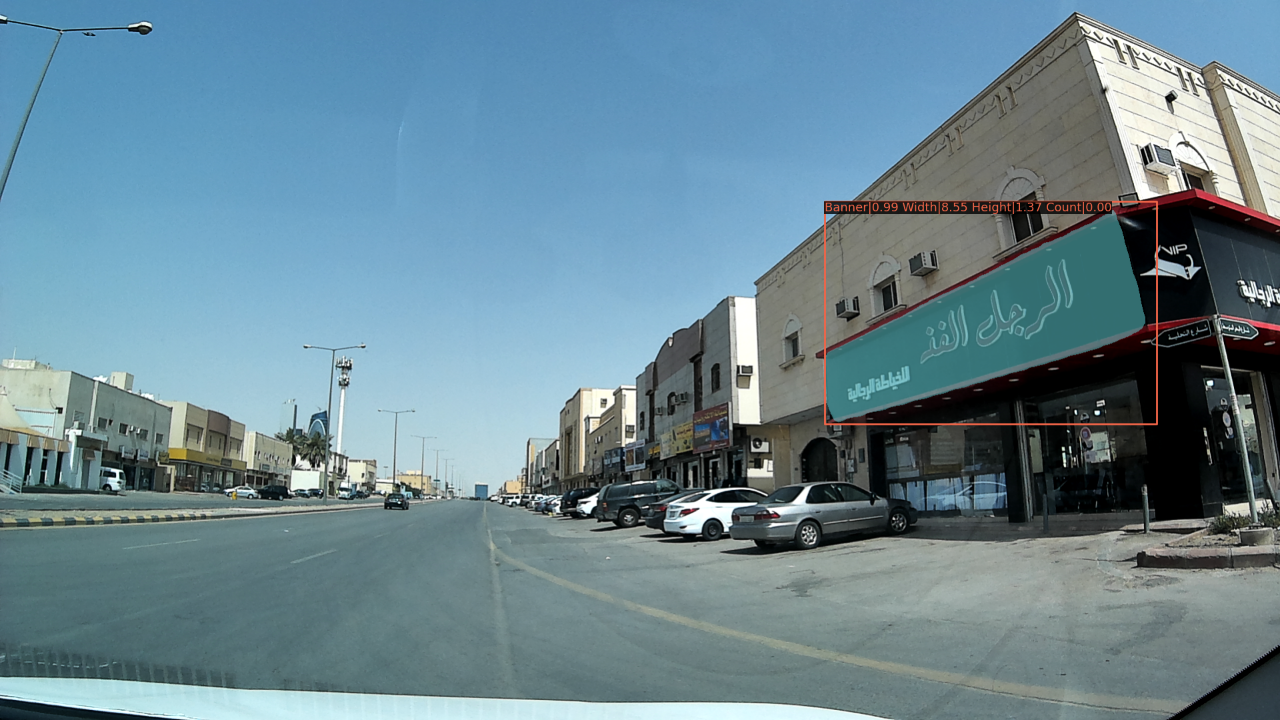}
\includegraphics[width=0.49\linewidth]{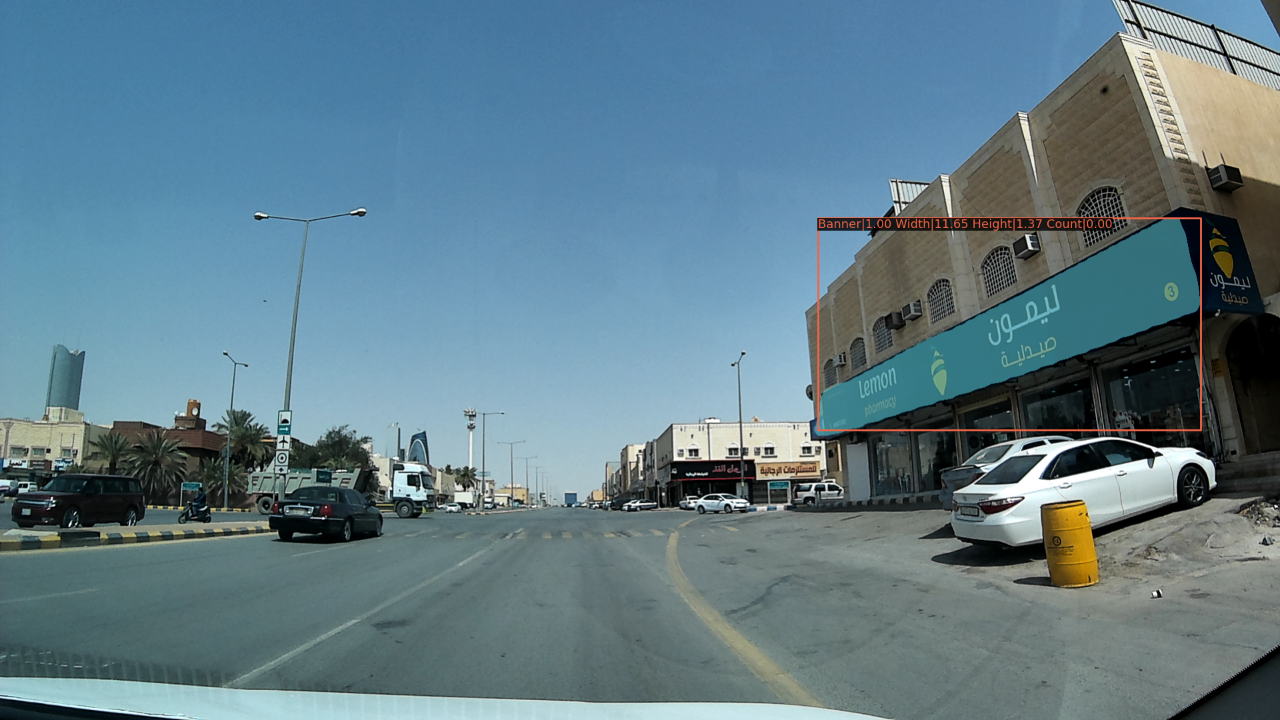}
\includegraphics[width=0.49\linewidth]{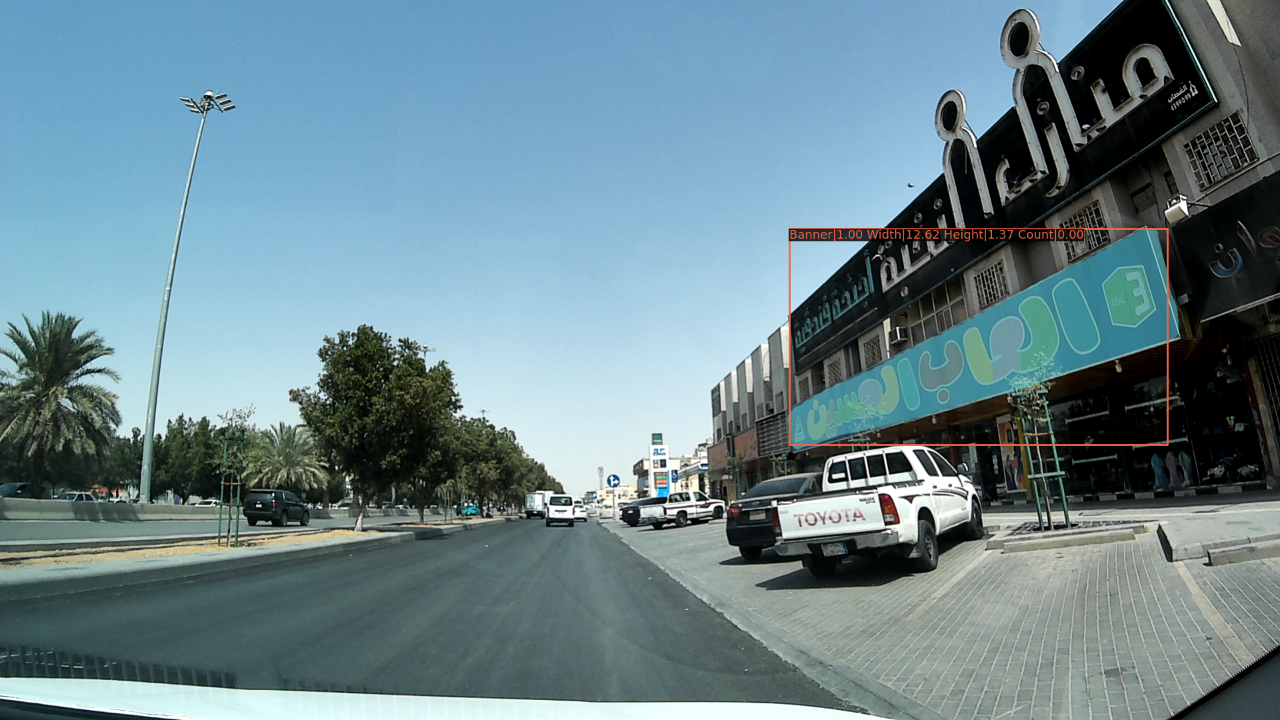}

\caption{Results of the PMODE on street shop signages. First row: The first result shows a width of 3.30 and a height of 1.23 meters. The second result shows a width of 8.55 and a height of 1.32 meters. Second row: The first result shows a width of 11.65 and a height of 1.37. The second result shows a width of 12.62 and a height of 1.37 meters }
\end{center}
\label{fig:intro}
\end{figure*}

The main contribution of our work is the dimension estimation component in the Neural Network architecture using the MLP that takes the mask coefficients from the detector head and, at the same time, applies them to the prototype mask generated from the protonet. The novelty of this work is that the network is trained to learn to look at the segmented pixels and predict the dimension of the shop signage object. The shop signage dimension is evaluated using a moving car setting with a camera fitted in the dash. The camera is directed towards the right side of the road, which is mainly populated with shops having signages on top of them.

During the evaluation, we also check the semantic consistency of the object dimension estimation. We define semantic consistency by the robustness of the model, wherein it can estimate the dimension of the same board with marginal deviation. It is done by making inference on multiple image frames of the same shop signage from the video and performing inference for nearby shop signages that can be measured in the subsequent frames.

In the remaining part of this paper, we present the related work for dimension estimation, shop signage dataset, and instance segmentation. We present the dataset details in the following section. We explain the details of our method by stating the approach, the overall architecture of the neural network, and the loss function details. We further elaborate on our experiments using various standard augmentations and custom augmentations, followed by a monocular depth estimation experiment. Finally, we submit our results on real-world test data and mention the future contribution to this work by incorporating a feedback system to improve the model performance using supplementary labeled data.
\begin{figure*}
\begin{center}
\includegraphics[width=0.33\linewidth]{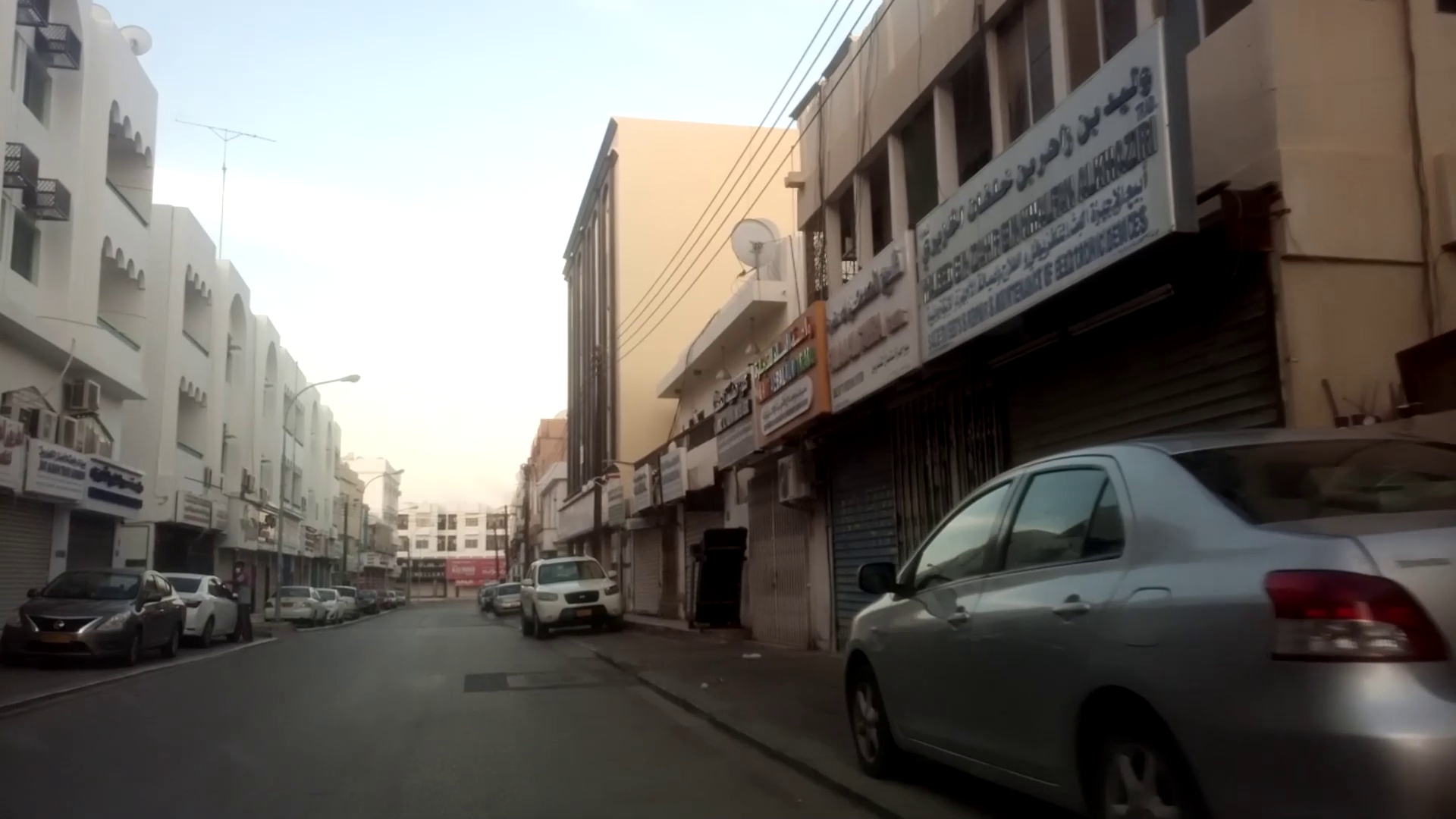}
\includegraphics[width=0.33\linewidth]{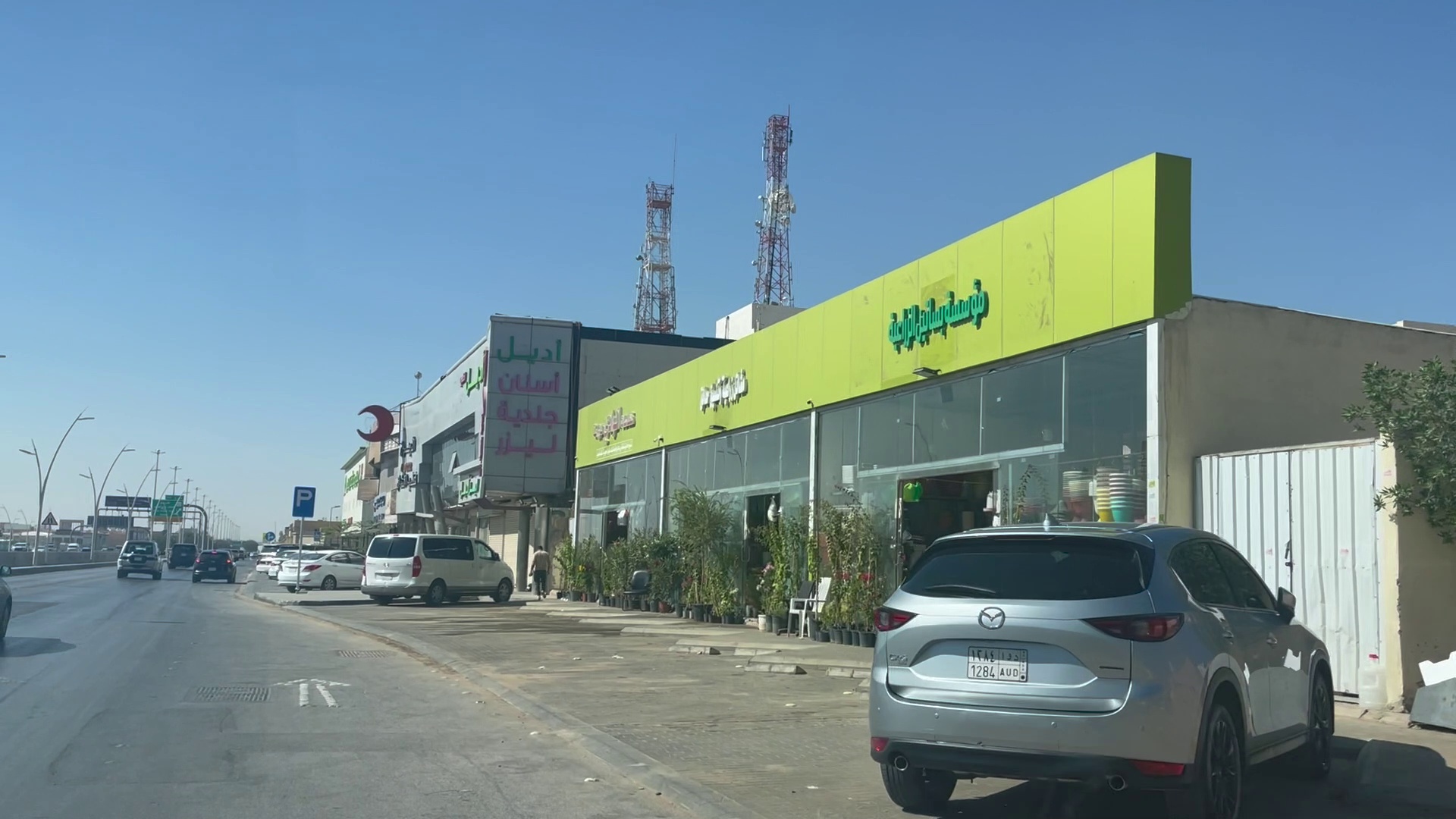}
\includegraphics[width=0.33\linewidth]{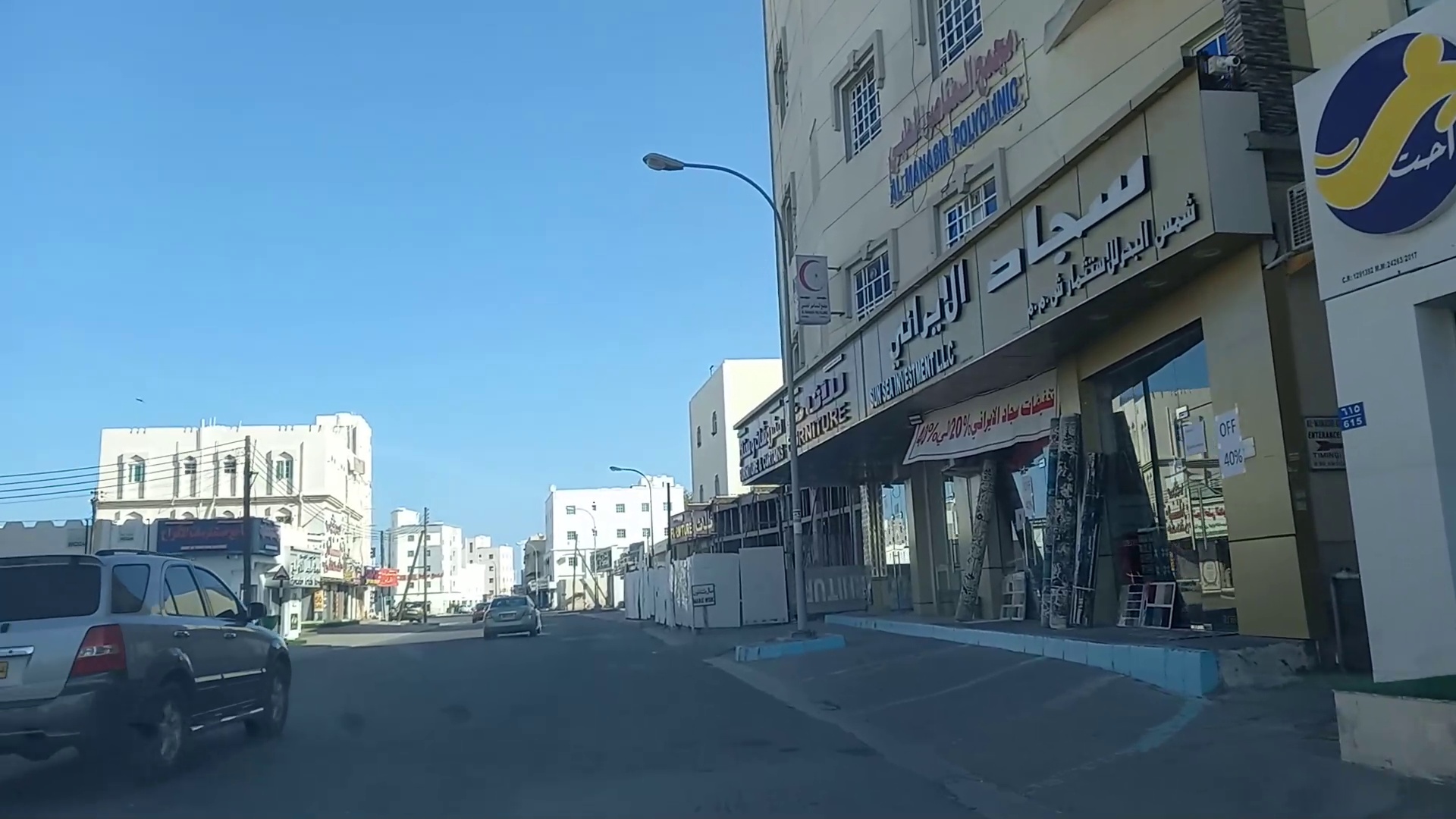}
\includegraphics[width=0.33\linewidth]{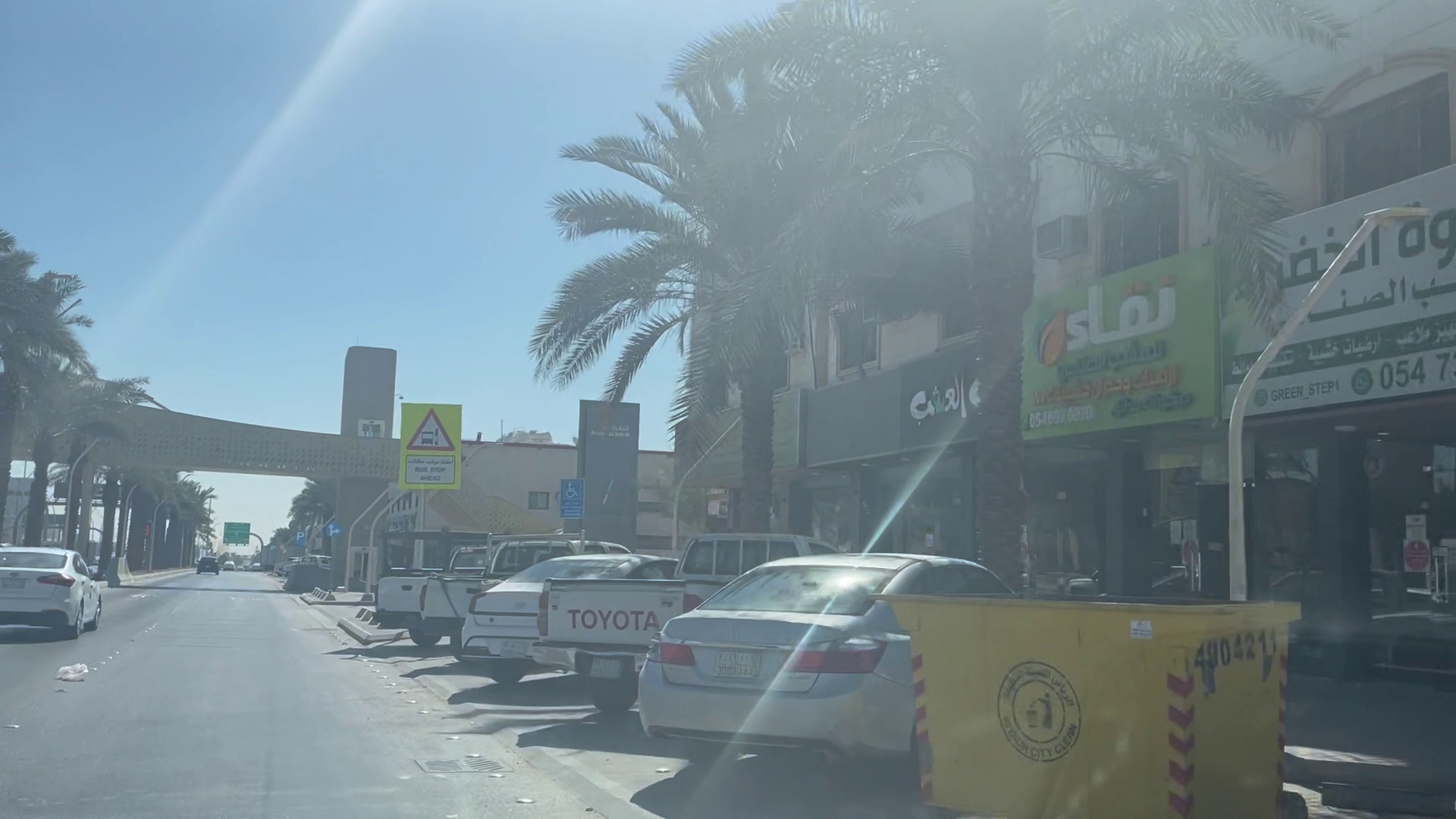}
\includegraphics[width=0.33\linewidth]{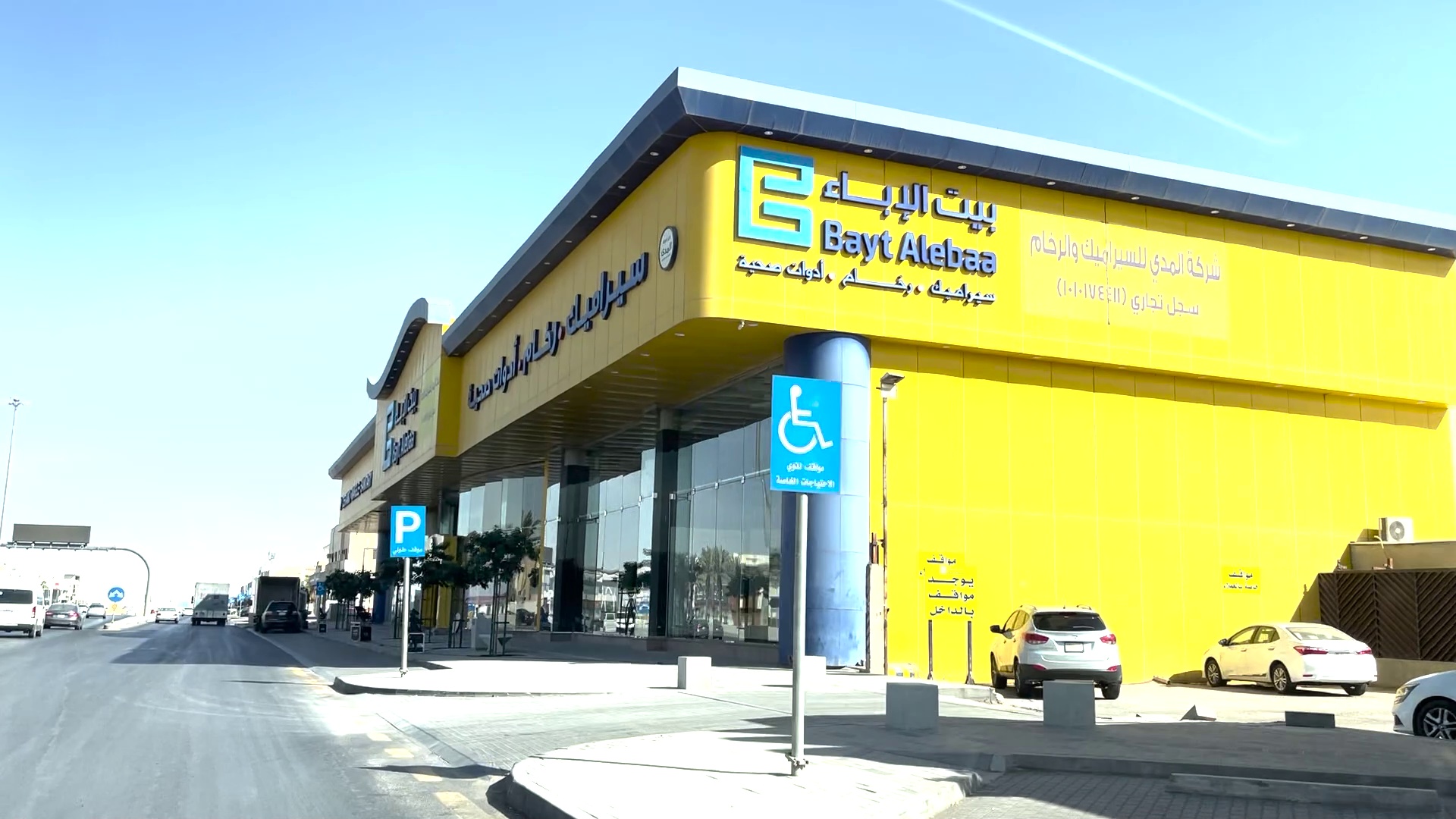}
\includegraphics[width=0.33\linewidth]{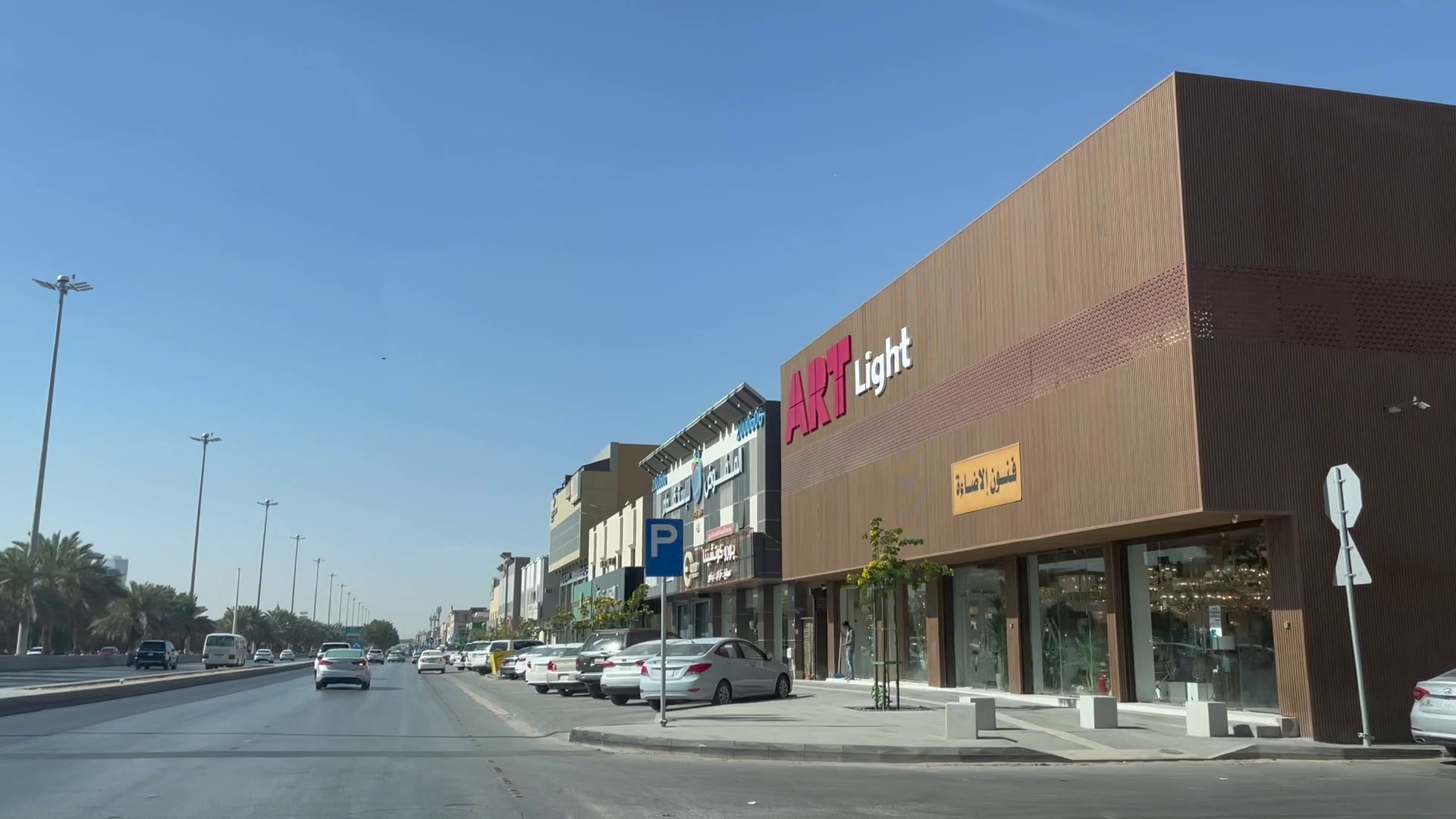}
\caption{Results of the PMODE. First row: The first image shows shop signage of average dimensions, which is very close to the camera. The second and third images show shop signages having the same color, making segmentation difficult. Second row: The first image shows shop signages with multiple occlusions. The second image shows shop signage which is very tall and wide. The third image shows a shop signage which is as big as the facade  }
\end{center}
\label{fig:dataset}
\end{figure*}
\section{Related Work}
\subsection{Shop Signage} ShopSign \cite{zhang2019shopsign} is a related dataset containing diverse Chinese shop sign images and their text; however, it doesn’t have the dimension information. The dataset helps in solving Chinese scene text recognition. Signboard Text \cite{do3988103signboardtext} is a recent and related work focusing on Latin and Vietnamese text recognition. Zhang et al. \cite{zhang2021character} is another work where they perform street view text spotting using segmentation to support autonomous driving.

\subsection{Instance Segmentation}
Mask-RCNN\cite{he2017mask} has been the predominantly used neural network architecture, for instance segmentation tasks in videos and images. As mentioned in \cite{cao2020sipmask}, it is a two-stage approach like other existing methods such as \cite{liu2018path, chen2019hybrid, huang2019mask, fang2019instaboost}. Mask-RCNN uses RoIAlign feature pooling and RPN(Region Proposal Network) to derive features for each region proposal. These features are then used for bounding box detection and segmentation mask prediction. Latest techniques such as Mask Scoring R-CNN\cite{huang2019mask} or Path Aggregation Network(PAN)\cite{liu2018path} needed re-pooling of the features and more computations. They become less relevant in real-time applications, which need 30fps or even more.

YOLACT\cite{bolya2019yolact}, on the other hand, is a real-time instance segmentation neural network that has proven strong segmentation results. The additional computational overhead after the on-stage backbone detector like ResNet-50 or ResNet-101 is minor for YOLACT. And especially during the evaluation time, it takes only approximately 5ms for the mask branch.SipMask\cite{cao2020sipmask} is a more recent instance segmentation neural network; however, the latency is more compared to YOLACT. A significant advantage of YOLACT is its generic nature, where the prototypes and mask coefficients can be applied to object detector networks. It should be noted that YOLACT has a Fast NMS (Non-Max Suppression) approach, which is 12ms faster than the traditional NMS. YOLACT also utilizes pixel-wise binary cross entropy loss during training which helps in the dimension estimation.

\subsection{Dimension Estimation}
In recent years, a lot of work has been done using Deep Learning for regression problems\cite{lathuiliere2019comprehensive} using computer vision. However, most of it has been on tasks such as pose estimation, facial landmark detection, and age estimation. Angela Tam et al. \cite{tam2003quadrilateral} is one of the earliest works done in quadrilateral signboard detection and text extraction. They use Hough transforms\cite{shapiro2001computer} and other image processing techniques to detect the standard quadrilateral signboards. Similarly, \cite{shapiro2001computer} makes use of OpenCV canny edge detection\cite{canny1986computational}, dilation, and erosion algorithms. \cite{vo2021automatic} makes use of a 3D camera for the depth, and apart from that, they have preprocessing, object detection, key points extraction, and depth interpolation before size calculation. \cite{kainz2015estimating} uses thresholding, canny edge detector\cite{canny1986computational}, Harris corner detector\cite{harris1988combined}, and mathematical estimations based on pixel values and reference objects. 

\section{Dataset}
We collected a novel dataset that comprises videos of the shop signs and their actual dimension measurements for this work. The dataset consists of 6500 shop signs with their dimensions. For a shop sign board with the dimension labeled with it, three frames contain it from different views. The dataset also contains shop signs that were taken during the night. The dataset includes shop sign boards that are incomplete and occluded due to the presence of trees and other objects on the street, like traffic signs.

\subsection{Data Collection}
The shop signage dimension dataset was collected from multiple countries in the Arab-speaking world. Hence the shop signages are filled with either Arabic shop names or English shop names and mostly a mix of both. The problem of dimension estimation and the solution is not limited to the region. The videos of shop signages were collected using mobile phones, which are kept horizontally on the windshield of a car. The videos were taken at Full HD resolution (1920x1080) and recorded at 30fps. 

The dimension of shop signage is defined as the width and height (in meters) of the frontal part of the shop signage. The depth of the shop signage is not included in the estimated dimension. In an image, the width and height comprise the physical quadrilateral board, which is kept on top of a shopfront to display the shop name correctly. During data collection, the dimensions of the shops were measured using laser tools with the support of tripods. At least two persons are required to take each shop sign's dimensions accurately.

Some shop signages are very tall. They are more than 3 meters in height, and some may even be as tall as the building façade. The dataset also contains certain shop signages in the videos whose dimensions are not recorded. Their dimensions could not be captured due to the limitations of the laser tool used during the process. During the manual data collection process, Shop signages that are black or blue do not reflect the laser properly.  
\begin{figure}[t]
\begin{center}
\includegraphics[width=0.49\linewidth]{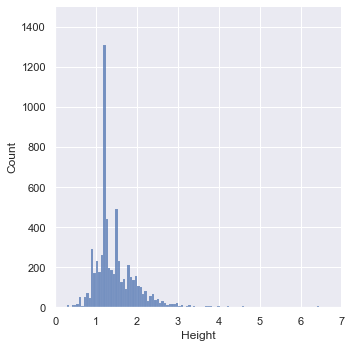}
\includegraphics[width=0.49\linewidth]{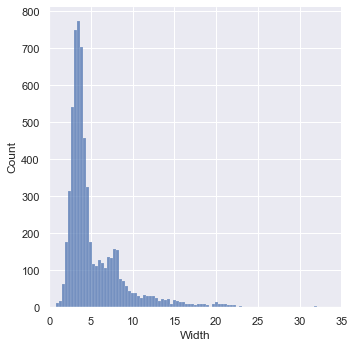}
\end{center}
   \caption{Distribution plots for Height and Width of our shop signage dataset}
\label{fig:long}
\label{fig:onecol}
\end{figure}
\subsubsection{Data Annotation}
Three nearest image frames from the video of each shop signage were annotated using polygon masks to get the correct quadrilateral. The right-most shop signage, which is fully visible, was annotated in each image frame. Faraway shop signages were not annotated because their borders would probably meet at the horizon. The dimension estimation of such signages would be highly error-prone. Partially visible shop signages were not annotated because they will not help estimate correct signage dimensions. Occluded shop signages are also annotated so that the model learns to detect and segment the signage regardless of the occlusion.
\section{Method}
\subsection{Approach}
\begin{figure*}
\begin{center}
\includegraphics[width= \linewidth]{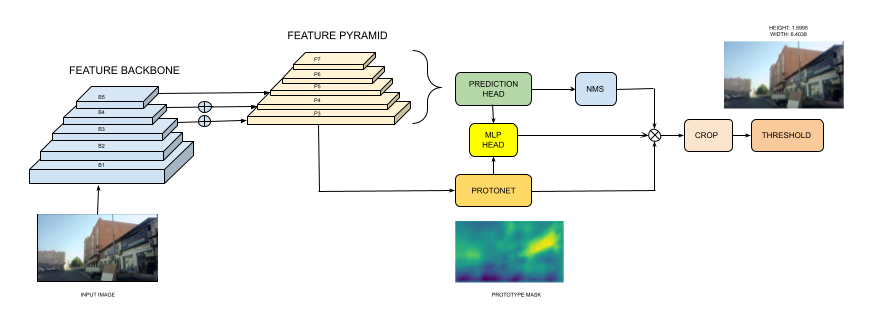}
\caption{The architecture of PMODE: The input image is passed to the feature backbone and then to the feature pyramid. Parallel branches of protonet and bounding box prediction heads are forked from the feature pyramid. The MLP plays the crucial role of detaching the prototype mask and taking the mask coefficients from the detector head to regress the dimensions of the object }
\end{center}
\label{fig:arch}
\end{figure*}
First, the object of interest, the shop signage, must be segmented using instance segmentation. Furthermore, only the nearest visible shop signage must be segmented to estimate the shop signage dimension individually and more accurately. The annotated images are combined with the width and height information from the excel sheet to form a final JSON file which contains the dimension information and mapping of the corresponding shop signage image. The images are resized from 1920x1080 to 500x500 for optimal computation of the feature maps using ResNet-50. The shop signages must be segmented first to get the correct quadrilateral pixels. Each shop signage should be of different segmentation pixel masks. Hence the right choice for architecture would be instance segmentation. 
\subsection{Overall Architecture}
YOLACT is a real-time one-stage instance segmentation neural network architecture. It has the advantage over other instance segmentation algorithms like MASK-RCNN in that it produces the object detection vector of mask coefficients for each anchor box and the prototype mask through two parallel branches. Once each instance goes through the NMS (Non-Max Suppression) with a high threshold, a mask is constructed by linearly combining the prototype branch with the mask coefficient branch. YOLACT has a faster NMS module than its predecessors, which is at least 12ms faster than the normal NMS. A sigmoid non-linearity follows this to produce the final masks. We use ResNet-50 as our default backbone feature extractor with a base image size of 500x500. Once the image features are extracted, it is sent to the Feature Pooling Network, which acts as the neck in the Neural Network. In our training configuration of YOLACT, there is a bounding box regression head with the object detection class with cross-entropy loss and a mask head for the prototypes with k channels of a maximum of 100 prototype masks to be trained. It also has a segmentation head that uses cross-entropy loss.
The Multi Layer Perceptron (MLP) eventually performs the signficant job of the dimension estimation. The protonet mask is detached from the Protonet along with the coefficients from the detector head to give input to the MLP head.
\subsection{Prototype Mask}
The prototype mask is one of the k prototype mask candidates obtained from the Protonet layer. The Fully Convolutional Network(FCN)\cite{long2015fully} generates it  parallel to the detector head.
We use the prototypes generated from the predictions using foreground and background prototypes for the segmentation loss.
\begin{figure}[H]
\centering
\includegraphics[scale=0.55]{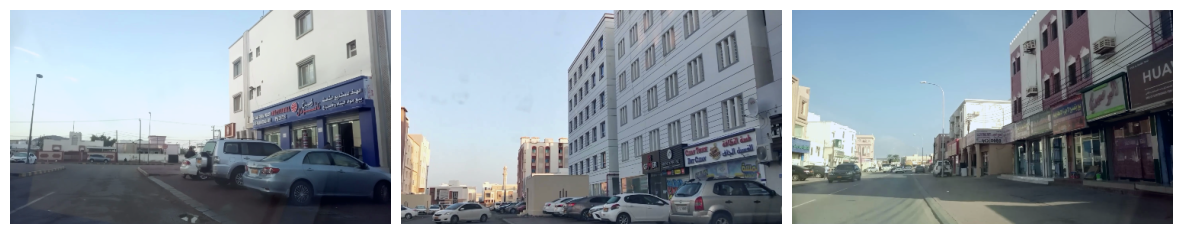}
\includegraphics[scale=0.55]{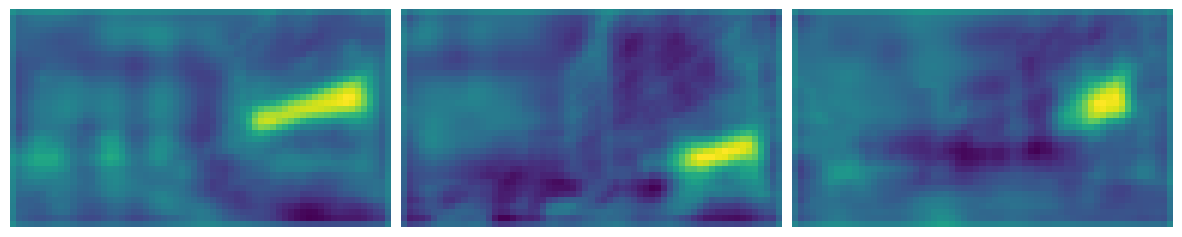}
\label{fig;prototype}
\caption{The first row shows the ground truth shop signages. The second row shows the prototype mask predictions for the corresponding images}
\end{figure}

Then we use the masks taken from different prototypes to calculate the mask loss using Binary Cross Entropy which is given by the equation
\begin{displaymath}
{L} = -\frac{1}{N}\sum_{i=1}^{N}y_{i}.log(p(y_{i})) + (1-y_{i}) . log(1-p(y_{i}))
\end{displaymath}

\subsection{Dimension Estimation}
The configuration mentioned in section 4.2 is enough for training a neural network with a dataset containing polygon masks of the shop signages. However, to do the regression of the height and width of the shop signages, we need to regress the two values in the correct head of the YOLACT network. We created a simple Multi-Layer Perceptron(MLP) with six layers of neurons that have Rectified Linear Units (ReLU) as the activation function. The input of the MLP is the mask of shop signage, and the regression output of this neural network will be two values representing the height and width dimensions. We selected the protonet head to add the loss of creating a prototype representative mask that can be used for regressing the height and width of the shop signage. We have empirically concluded that the loss value converges well when the mask size is 150x150. This loss is referred to as hnw-loss $L_{hnw}$. It is further aggregated to the prototype mask loss.

\subsection{Loss Function}

The total loss of the PMODE network consists of terms corresponding to segmentation mask generation, bounding-box detection (classification and regression), and height and width regression. For mask generation, we use the Binary Cross(BCE) Entropy loss $L_{seg}$ mentioned in section 4.3. This will help the dimension estimation since YOLACT looks at all pixels. During training, we use the smooth L1 loss for the height and width regression of $L_{hnw}$. The classification loss $L_{cls}$ is Cross Entropy loss, and the bounding box detection loss, $L_{bbox}$, is Smooth L1 loss.

\[ L_{total}=L_{seg} + L_{hnw} + L_{bbox} + L_{cls}\]

\section{Experiments}
We prototyped the solution on NVIDIA 3070 RTX and 3090 RTX computers and further developed the high-performance models on NVIDIA DGX-1 with 8 V100 GPUs. We conducted the experiments with various shuffle configurations. Since the dataset was collected from two countries, we did the training and validation with a mix of shop signage images from these countries. The dataset contains 5000 shop signage images and their dimensions from one country and 1300 shop signages and their respective dimensions from the other country. All experimentation has been done using the MMdetection\cite{chen2019mmdetection} library. 
\subsection{Augmentations}
We experimented with light and heavy data augmentation using Albumentations\cite{buslaev2020albumentations}. The results provided an extra performance on rainy and dusty days. The augmentations techniques include photometric distortion, random flips horizontally and vertically, slight rotation up to 15 degrees, random brightness contrast, RGB shift, hue saturation value, jpeg compression, channel shuffle, and median blur, as shown in the image below.

We made custom augmentation related to the methodology to improve the robustness of height and width estimations; such augmentation includes object mask length extension and 90 degrees rotation, flipping the ground truth height and width of the object of interest. The tests resulted in more robust estimations.
\begin{figure}[H]
\centering
\includegraphics[scale=0.35]{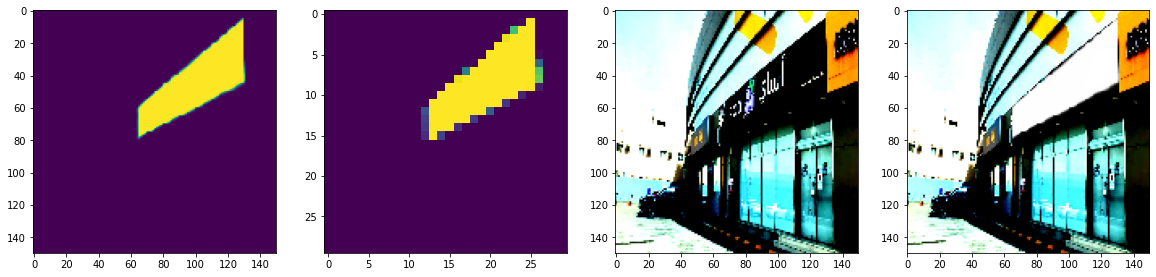}
\caption{Mask prediction of the Protonet, producing multiple masks that look similar to the left one, resized into 30x30 before inserting it into the MLP for dimension estimation. The data-augmented ground truth image can be seen on the right side.}
\label{fig;mask}
\end{figure}
\subsection{Depth Estimation}
We trained the estimator as embedded depth estimated masks using a monocular camera with MiDas\cite{ranftl2020towards}. The system shows robustness against further away shops while showing similar performance for closer shops. The depth estimation was integrated over the mask before the dimension estimation, as shown in the figure below. The depth augmentation and the corner loss mentioned in section 5.3 produced the best segmentation mAP of 0.720 and Height and Width MAPE of 0.223, as seen in Table 1. It is pretty intuitive that the depth feature adds a lot of value to any object dimension estimation. However, we are limited by the constraint of a monocular camera and the setting of a moving car.
\begin{figure}[H]
\centering
\includegraphics[scale=0.4]{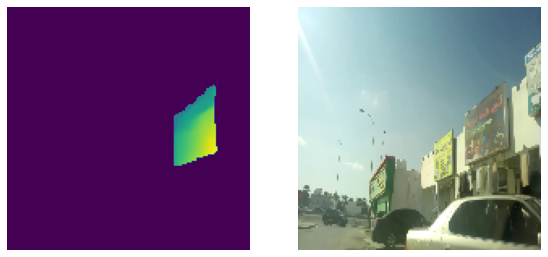}
\caption{Left image shows the predicted mask with depth estimation. While the right side shows the raw image}
\label{fig;depth}
\end{figure}
\subsection{Corner Loss}
While testing the height and width estimations, we noticed that while we are accounting for pixel prediction confidence, there is still the issue of the model predicting artifacts interfering with the height and width estimated values. So we made the corner loss as an extra loss embedded with the cross entropy to focus more on vertices, allowing height and width estimated values to be more accurate by concentrating more on segmentation shape than simply activating most pixels with the mask prediction. The corner loss starts by identifying corner positions in the ground truth label using the Harris Corner detection algorithm\cite{harris1988combined}. Then we proceed to cluster the corners using radius-based clustering algorithm DBSCAN\cite{ester1996density} as shown below in the figure:
\begin{figure}[H]
\centering
\includegraphics[scale=0.7]{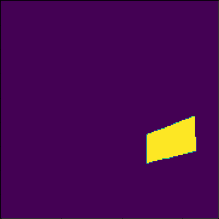}
\includegraphics[scale=0.7]{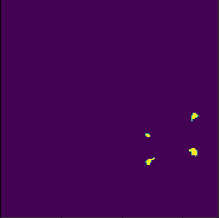}
\includegraphics[scale=0.7]{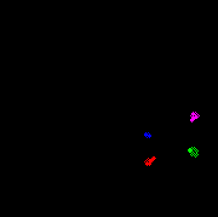}
\caption{Left image identifies the ground truth mask for a shop sign, the mid is showing after corner detection. while the right one shows the corner groups using the clustering algorithm}
\label{fig;corners}
\end{figure}
After that, we calculate the centroid of each clustered group; we use that as the center of the circle used in the final step to estimate the corner alignment between ground truth and mask prediction. We then proceed to estimate the shape edge using the canny edge algorithm
\begin{figure}[H]
\centering
\includegraphics[scale=0.5]{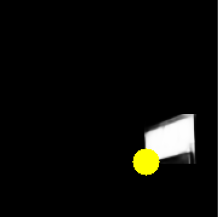}
\includegraphics[scale=0.5]{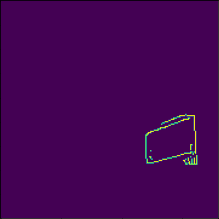}
\caption{Left image: The white rectangle represents the shop sign mask, while the yellow circle represents the radius of the center calculation. Right image: The shape contour after applying the canny edge algorithm}
\label{fig;corners1}
\end{figure}

We then use the circular radius over the edge detected mask to measure the center of the ground truth and for each corner for both predicted and ground truth. We then calculate the (1-r2) score for the position of both as our regression loss. If we faced the issue of the predicted mask not having active pixels in the radius, we proceeded with a maximum loss of 1 for the cluster group while accounting for the predicted pixels within the radius for the high loss.
During the tests done. The corner loss has resulted in around 0.1 mAP increase over the shop signage data and increased visual and dimension estimation accuracy over the test data.
The overall loss would then be re-written as 
\[ L_{total}=L_{seg} + L_{corner} + L_{hnw} + L_{bbox} + L_{cls}\]
where 
    \[L_{corner}= 
\begin{cases}
    1 - {R^2},        & \text{if $mask(x,y) == 1$ and $(x,y) \subset Corners$}\\
    1,              & \text{$otherwise$}
\end{cases}\]
$mask(x,y) == 1$ refers to the $(x,y)$ coordinate pixel being part of the shop signage prediction and $Corners$ refer to the set of all corners

\section{Results}
We trained the network for under 100 epochs with the dataset from both countries and observed that the dimension estimation converges ideally at the 21st epoch. As shown in Table 1, we find out the mean Average Precision for the bounding box loss, segmentation loss, and Height and Width loss using MAPE(Mean Absolute Percentage Error). The rows represent various training experiments with only Protonet Mask, the experiment with the protonet mask and DDepth the experiment with the protonet mask, and Depth with Corners loss and augmentations as explained in the Experiments section.  

\begin{table}[!htbp]
  \scriptsize 
  \centering
    \caption{Comparison of various experiments for shop signage dimension estimation.}
    \hspace*{-1cm} 
    \begin{tabularx}{0.8\linewidth}{Xlllllllll|l}
    \toprule
    {\sc Method}  & Bounding Box mAP 0.5:0.95  & Segmentation mAP 0.5:0.095 & Height and Width MAPE \\
    \midrule
    Protonet Mask & 0.634 & 0.717 & 0.235 \\
    With Depth & \textbf{0.645} & 0.645 & 0.238  \\
    Depth Aug. and Corner Loss  & 0.642 & \textbf{0.720} & \textbf{0.223} \\
    \bottomrule
    \end{tabularx}%
  \label{tab5_3}%
\end{table}%

The mask experiment is wherein we look at the pixel confidence and perform the dimension estimation. In Figure 10, we show the training loss with Smooth L1 loss for the Height and Width regression, whereas the validation loss is L1. The third graph shows the IOU values saturation for the bounding box and segmentation from the 21st epoch while keeping the mAP value 0.5:0.95. The Height and Width Validation is done using Mean Absolute Percentage Error(MAPE).
\begin{figure}[H]
\includegraphics[width=0.49\linewidth]{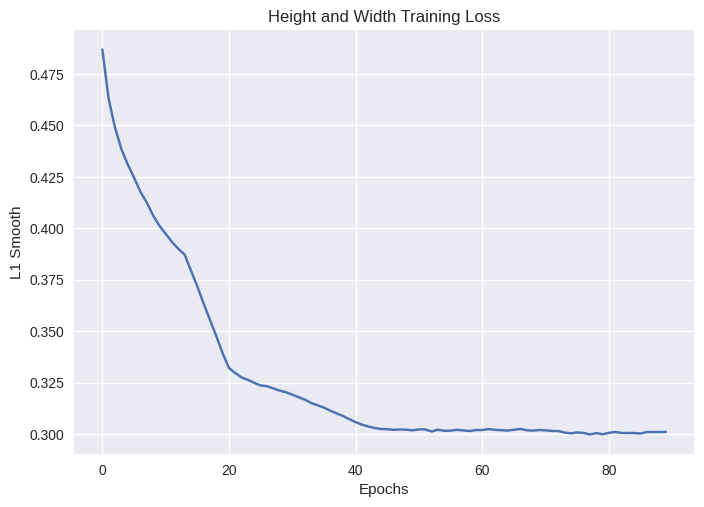}
\includegraphics[width=0.49\linewidth]{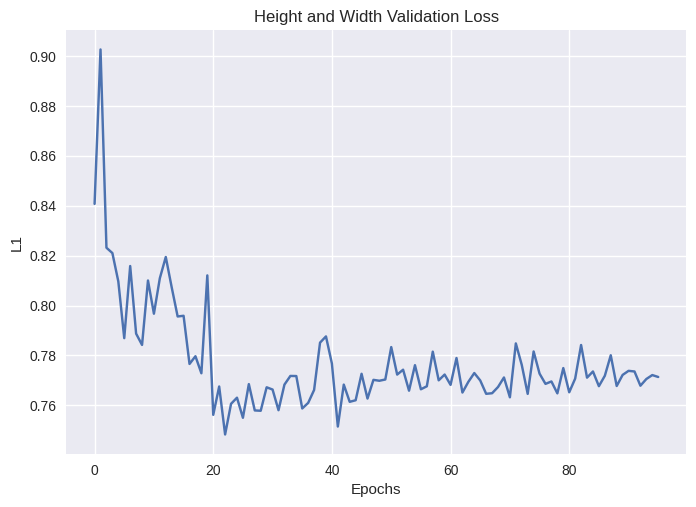}
\includegraphics[width=0.49\linewidth]{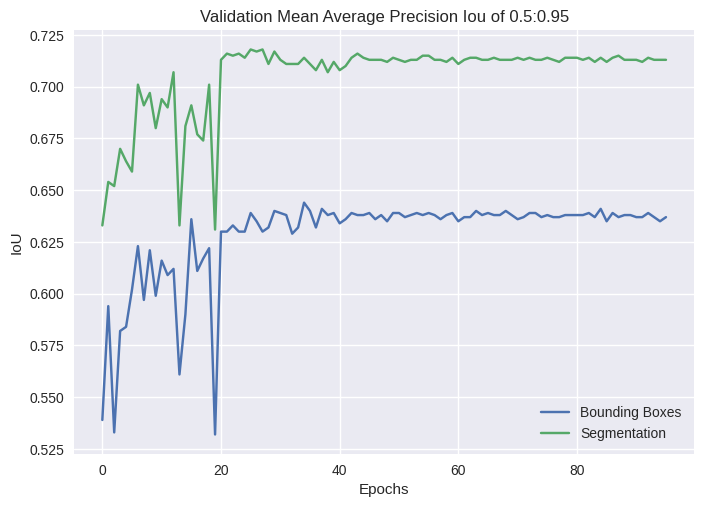}
\includegraphics[width=0.49\linewidth]{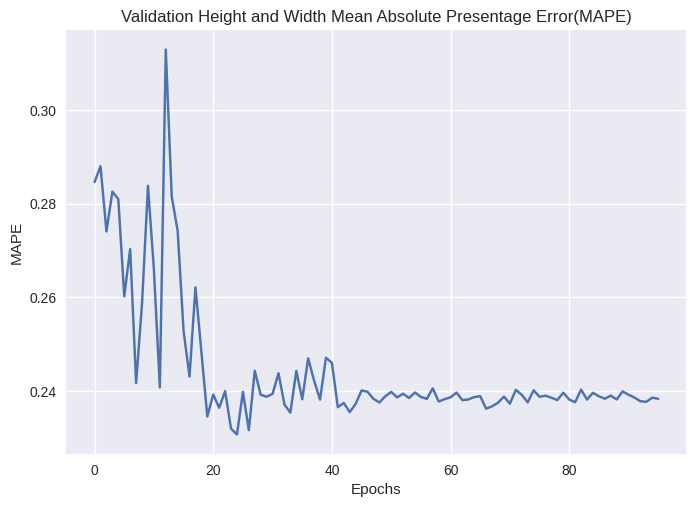}
\caption{The plots show the training and validation loss for the Depth augmentation with corner loss experiment for 95 epochs}
\end{figure}
\subsection{Semantic consistency in video}
We evaluated real-world scenarios to validate the real-time performance of the model and to check how the estimation is across objects that are comparable in size. A moving car equipped with a camera goes by the street with the shop signage on the right side of the road. The camera takes in the video stream at 30 fps. As shown in figure 11, when PMODE looks at the same shop signage, the dimensions are consistent with trivial variation in the dimensions. 
When the car moves forward to look at the next shop signages, as depicted in the further rows, the dimensions change according to the variation in the pixel information obtained from the prototype masks of PMODE. We observed that the model is quite robust in estimating the dimensions, given that the angle and orientation of the camera are kept intact.

\begin{figure}[H]
\includegraphics[width=0.33\linewidth]{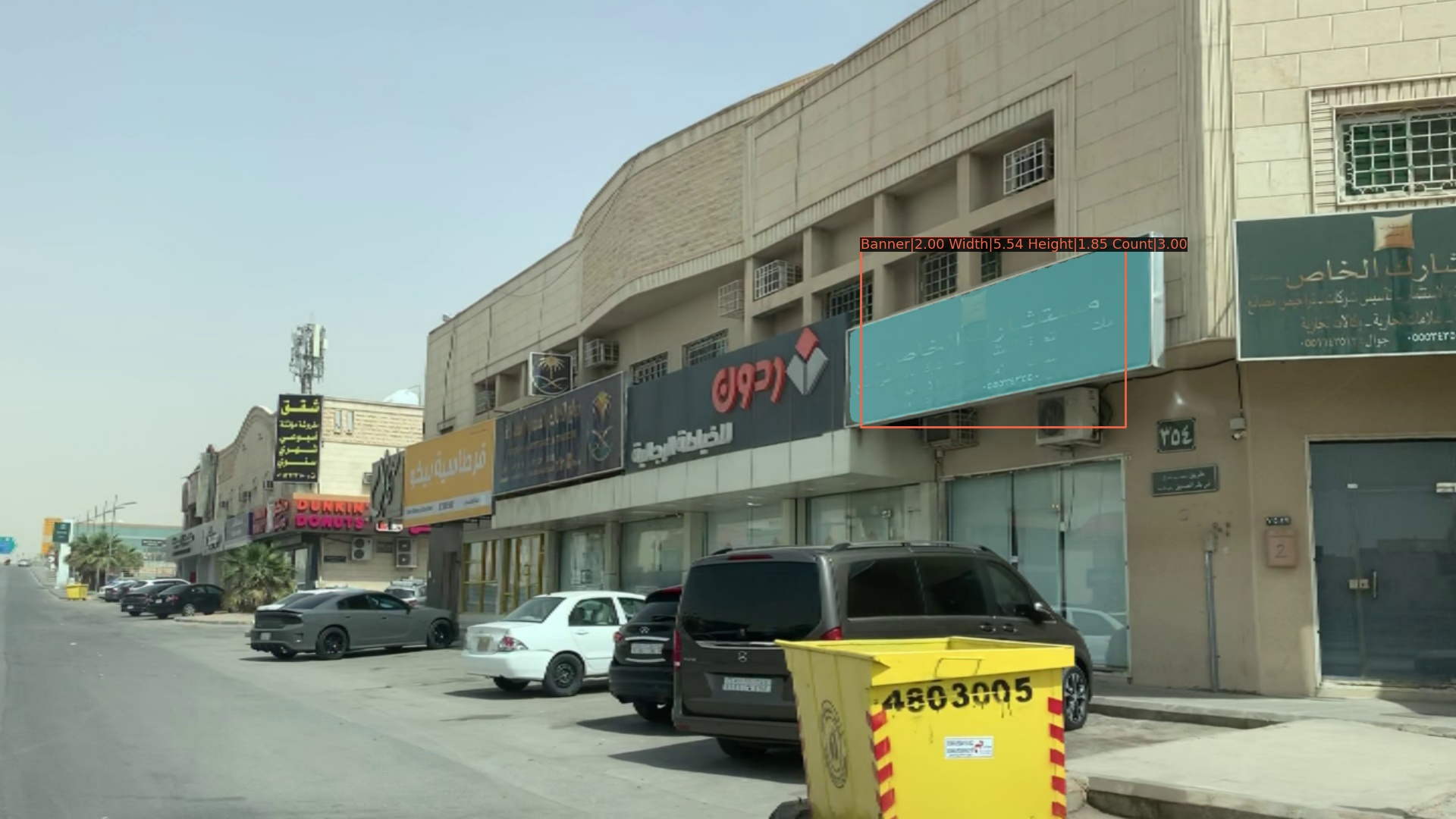}
\includegraphics[width=0.33\linewidth]{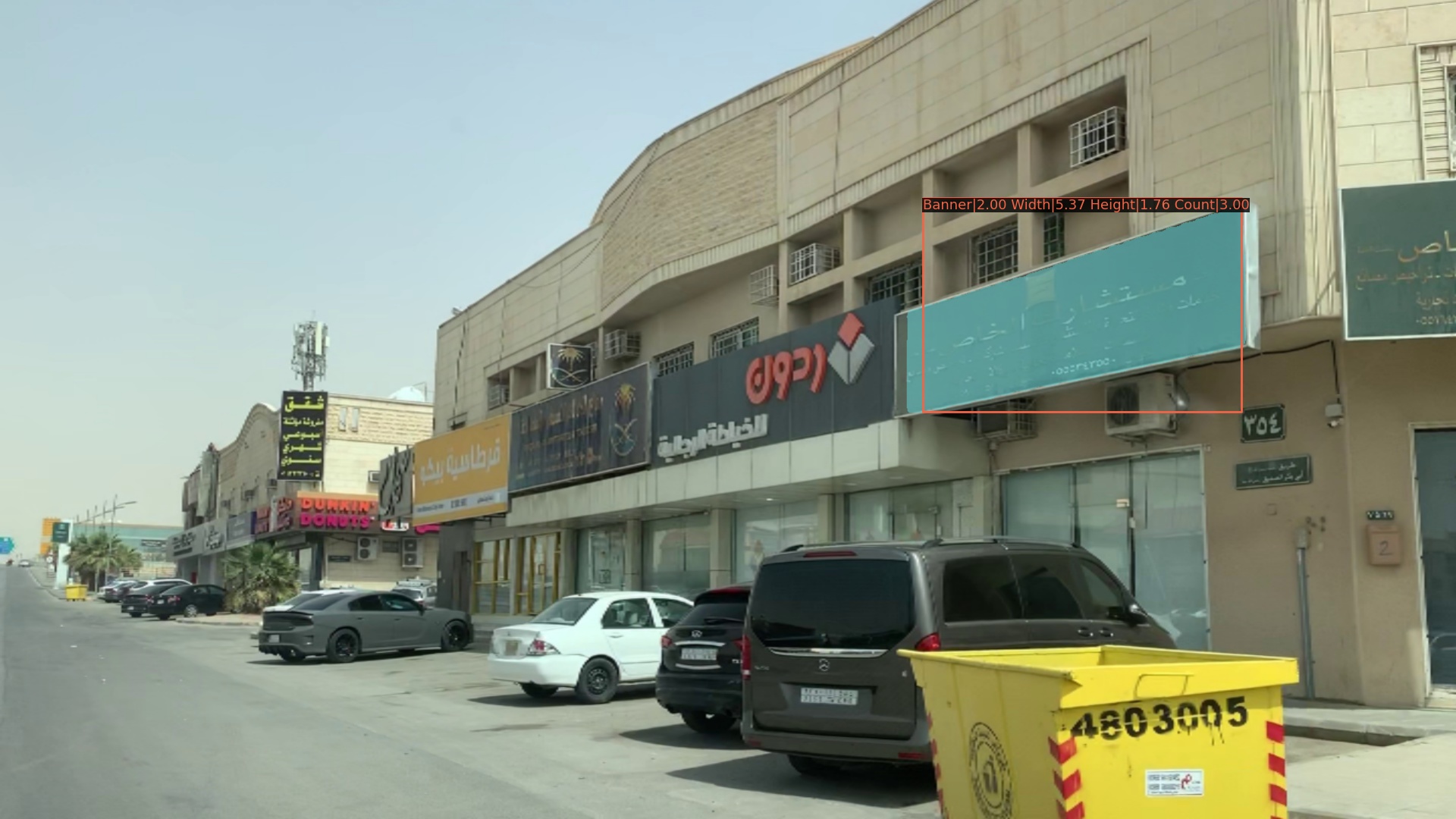}
\includegraphics[width=0.33\linewidth]{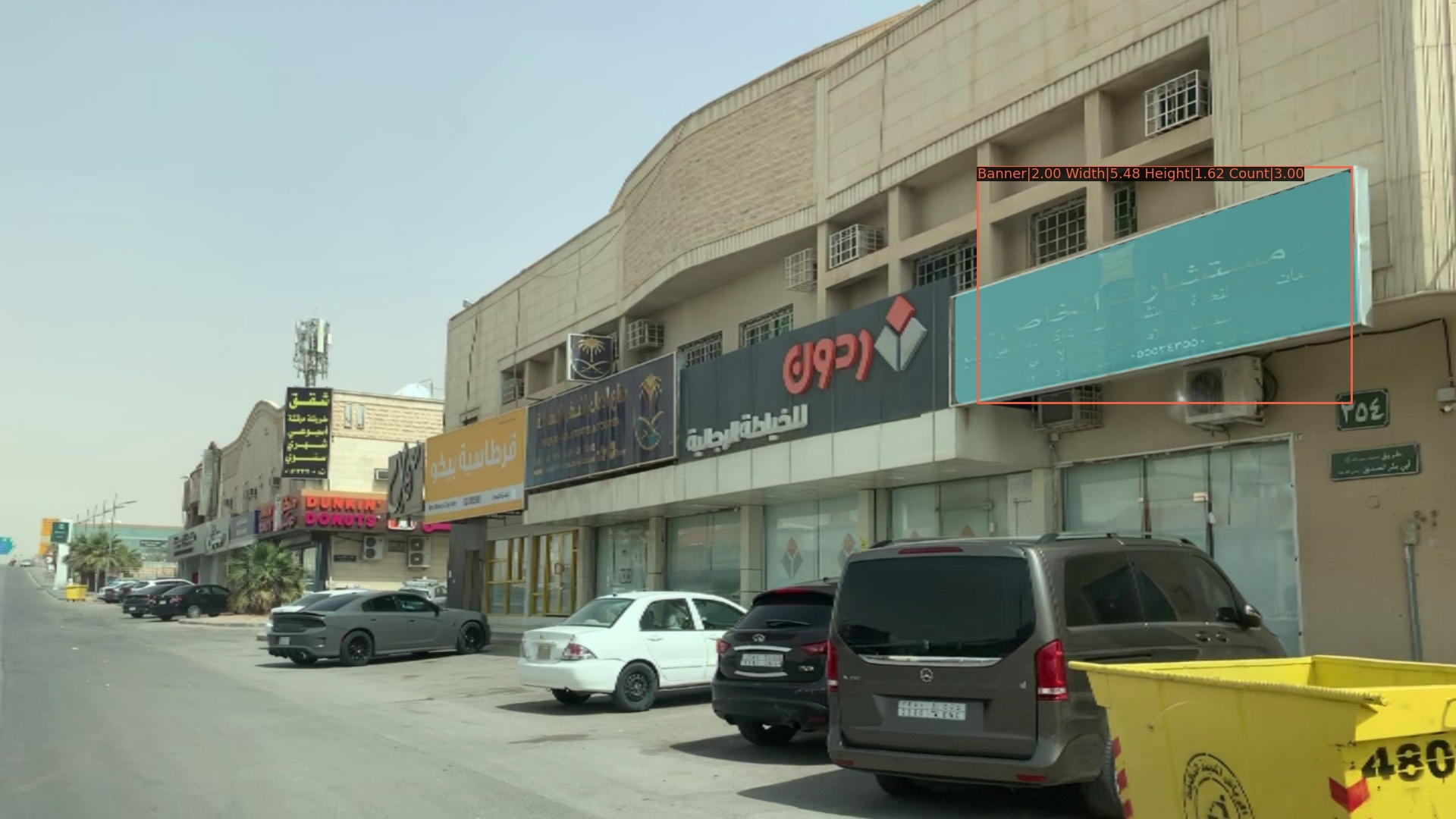}

\includegraphics[width=0.33\linewidth]{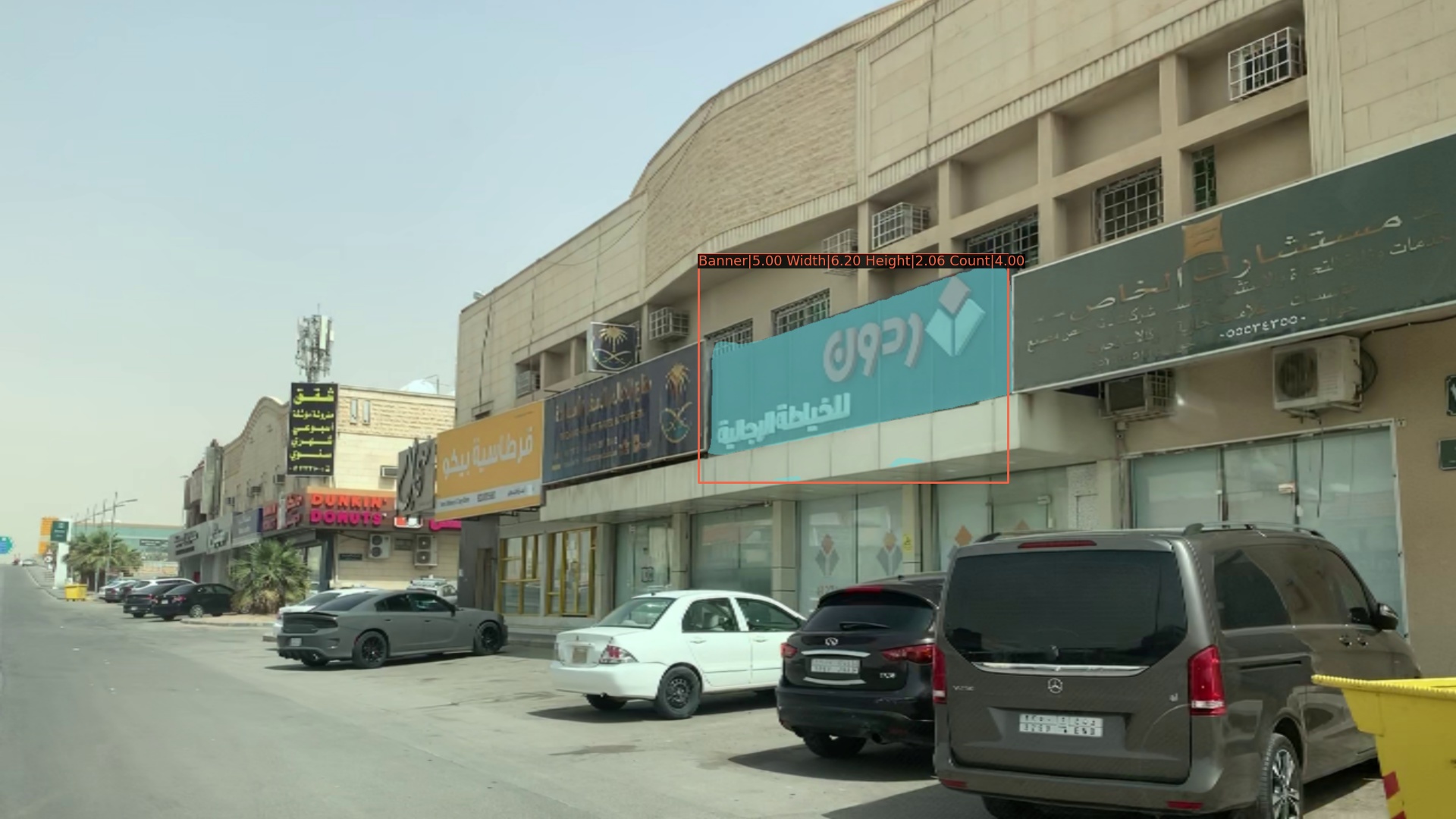}
\includegraphics[width=0.33\linewidth]{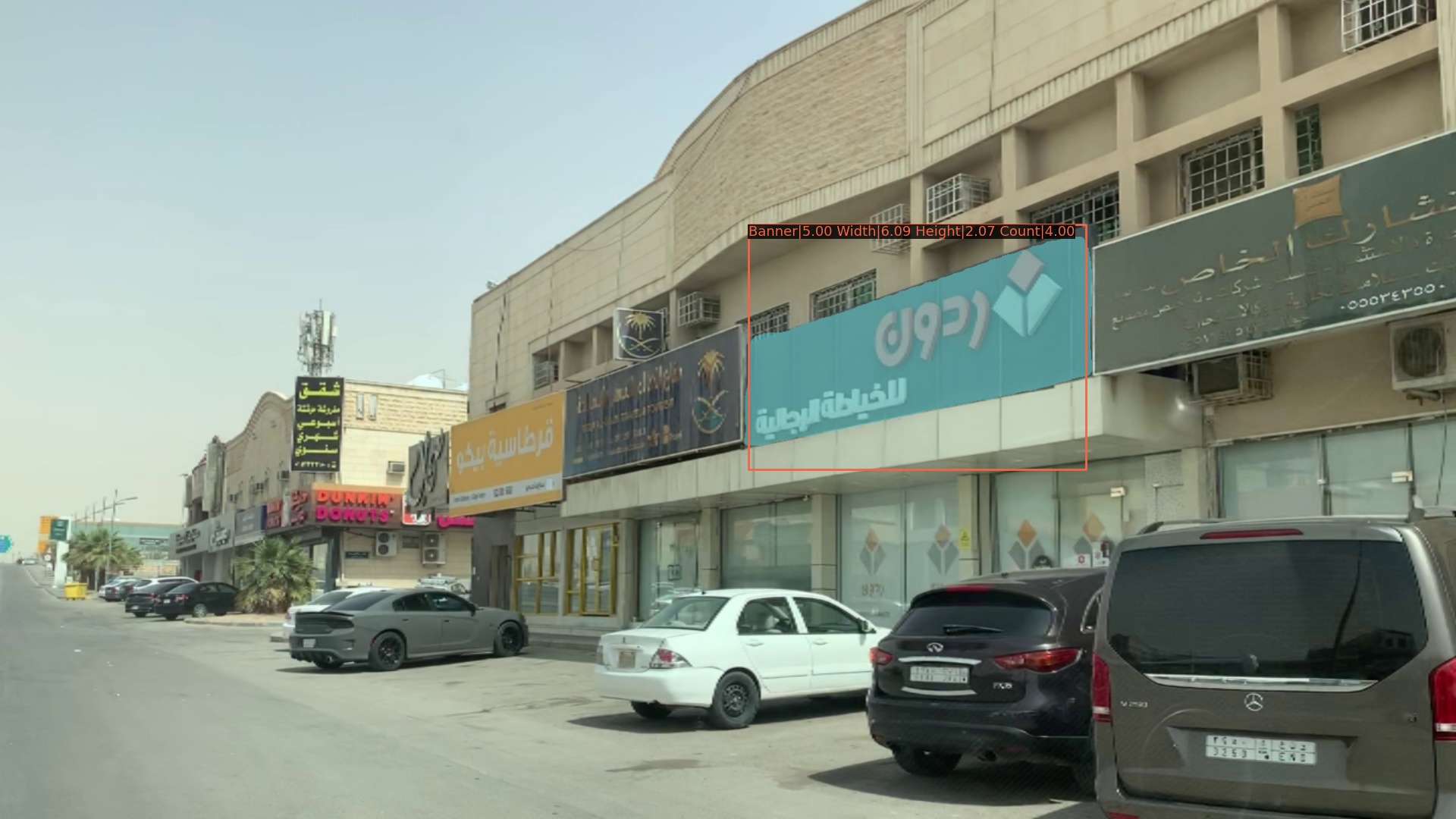}
\includegraphics[width=0.33\linewidth]{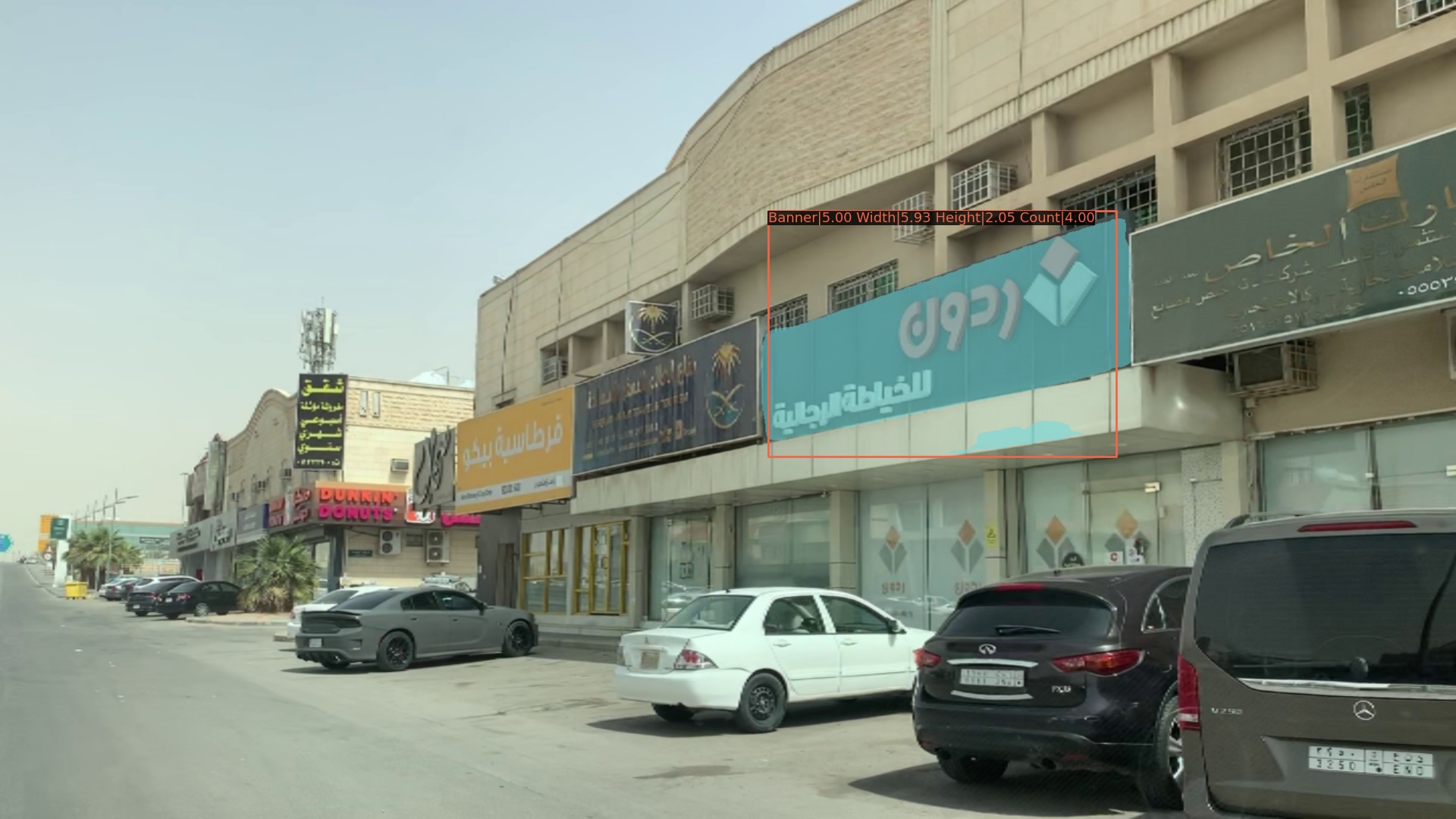}

\includegraphics[width=0.33\linewidth]{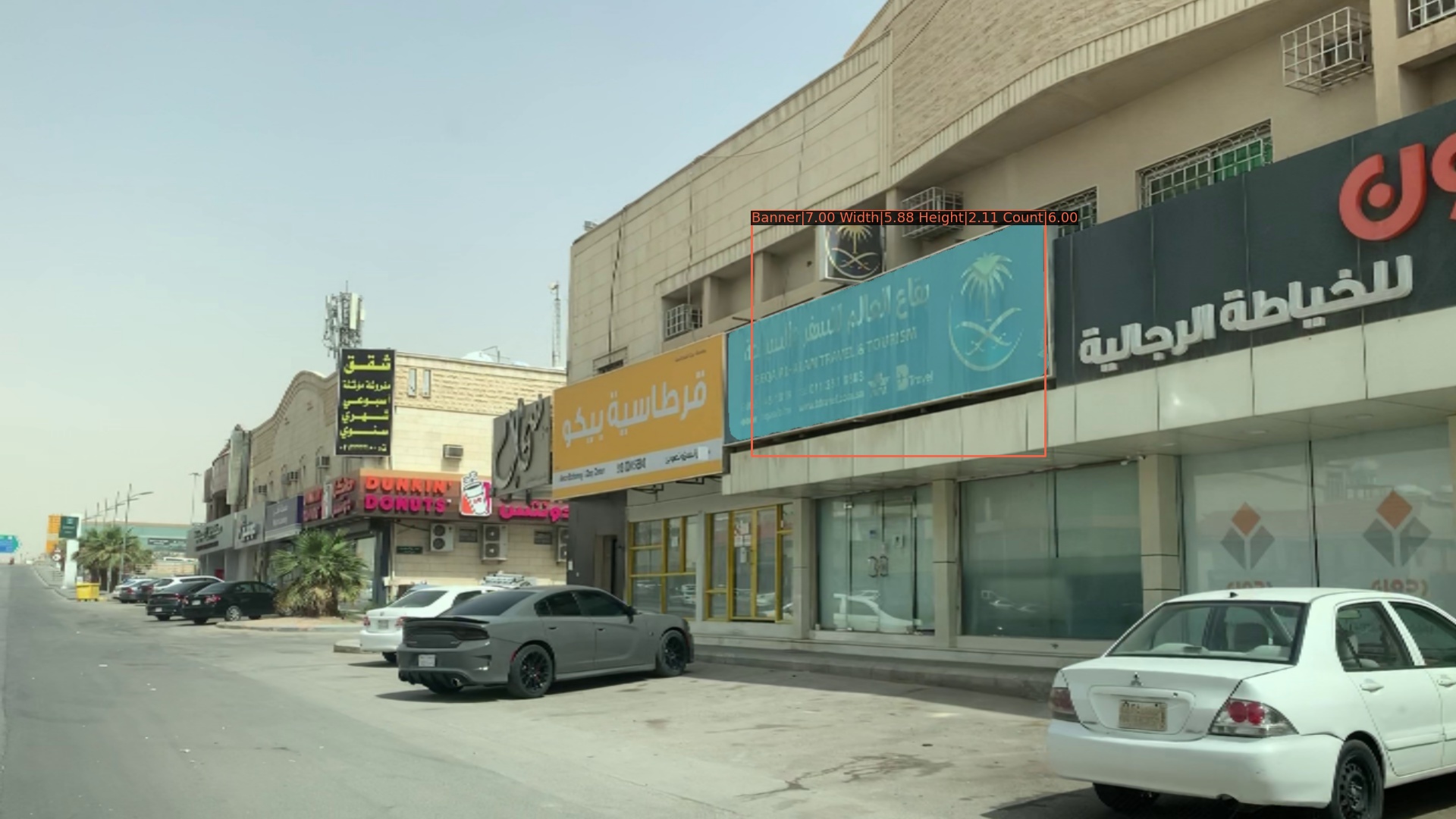}
\includegraphics[width=0.33\linewidth]{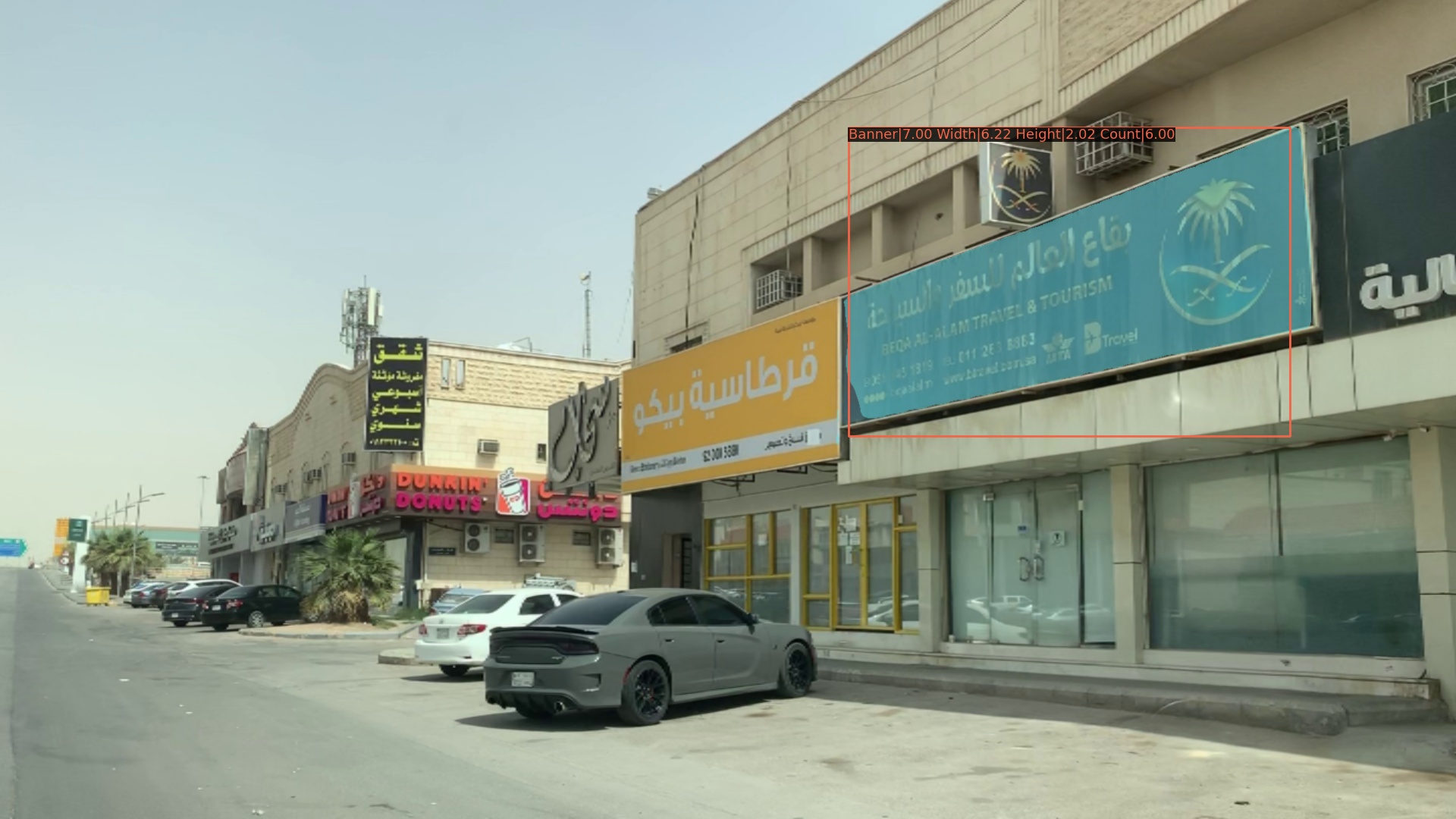}
\includegraphics[width=0.33\linewidth]{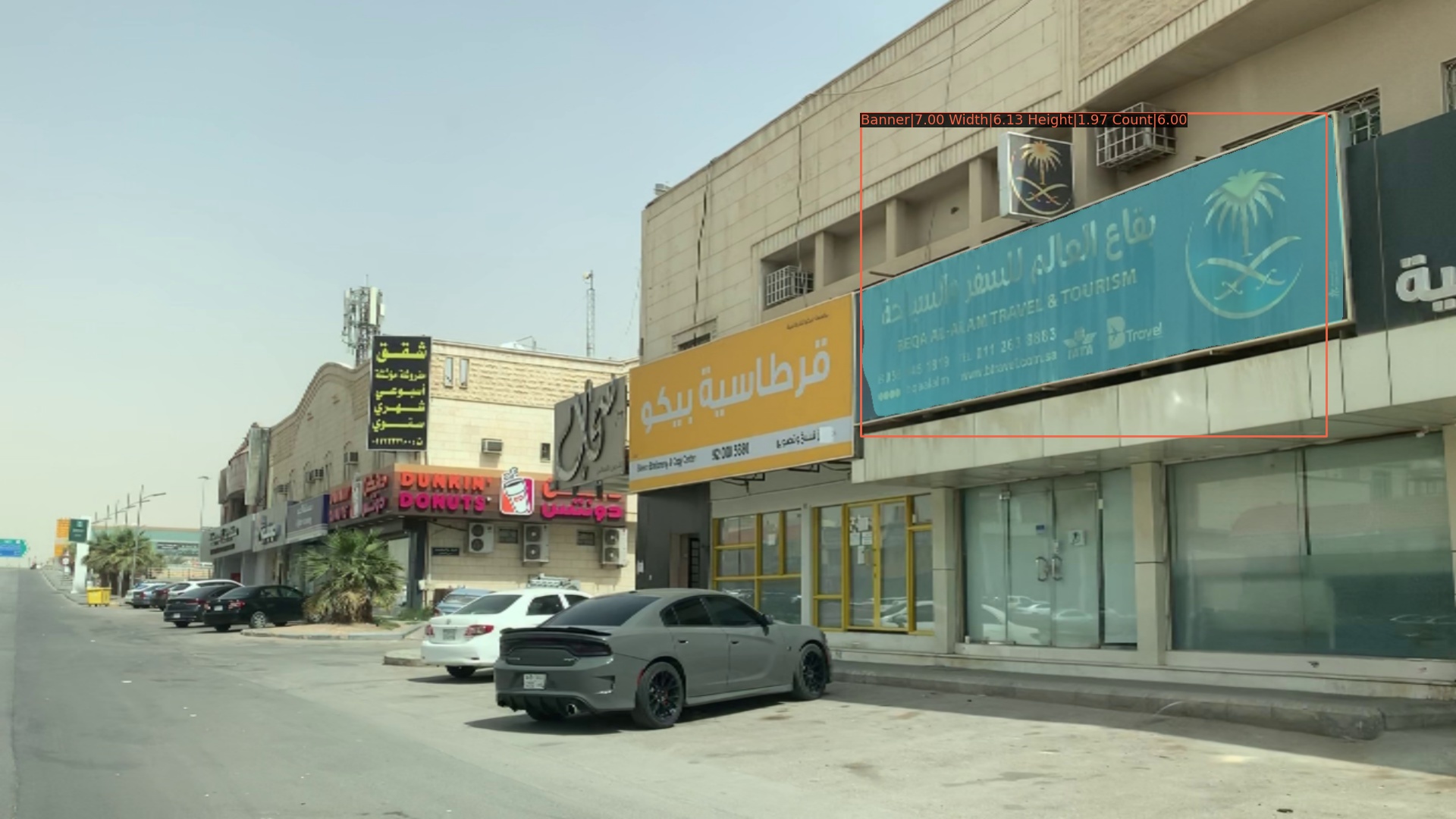}

\includegraphics[width=0.33\linewidth]{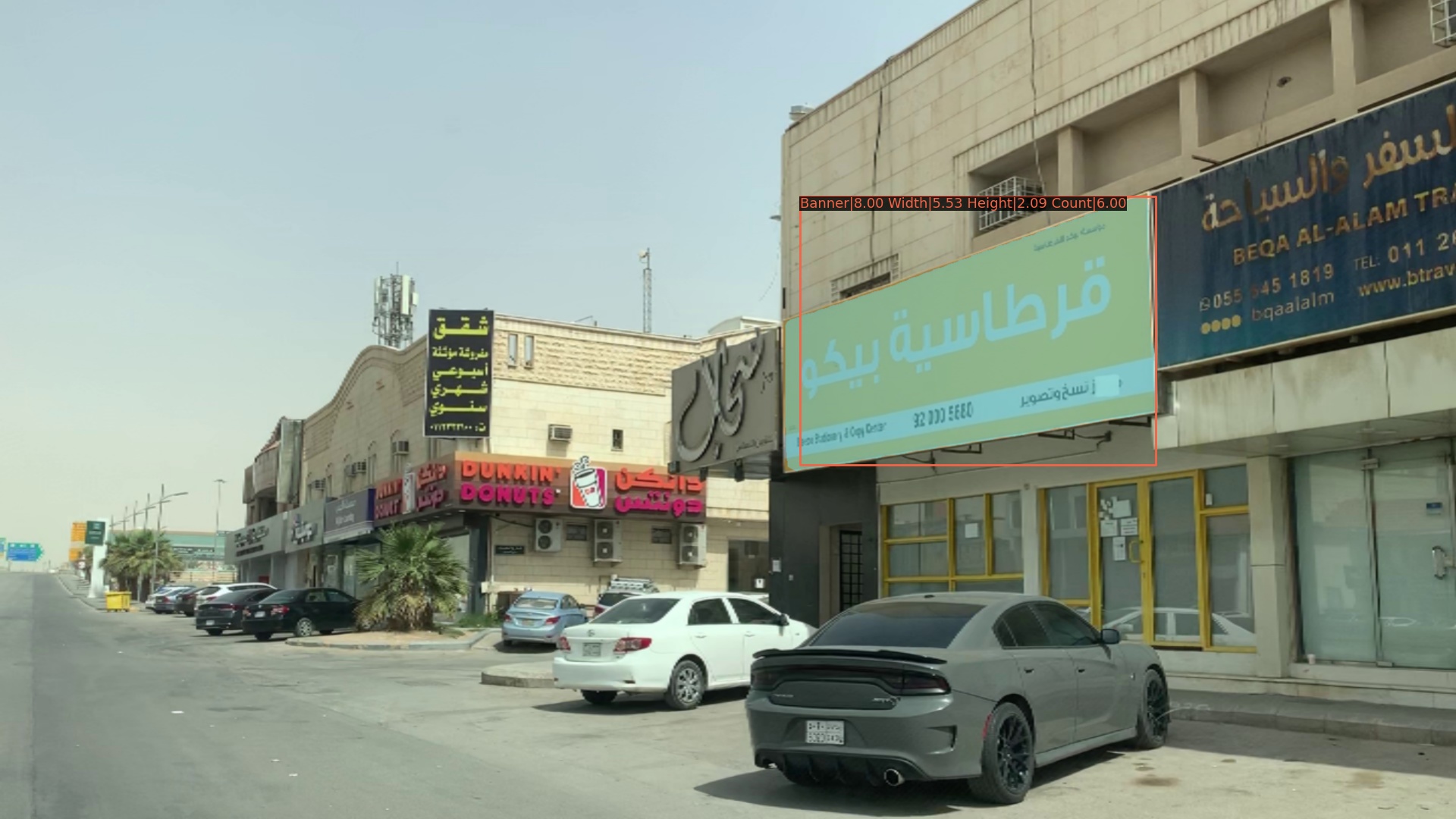}
\includegraphics[width=0.33\linewidth]{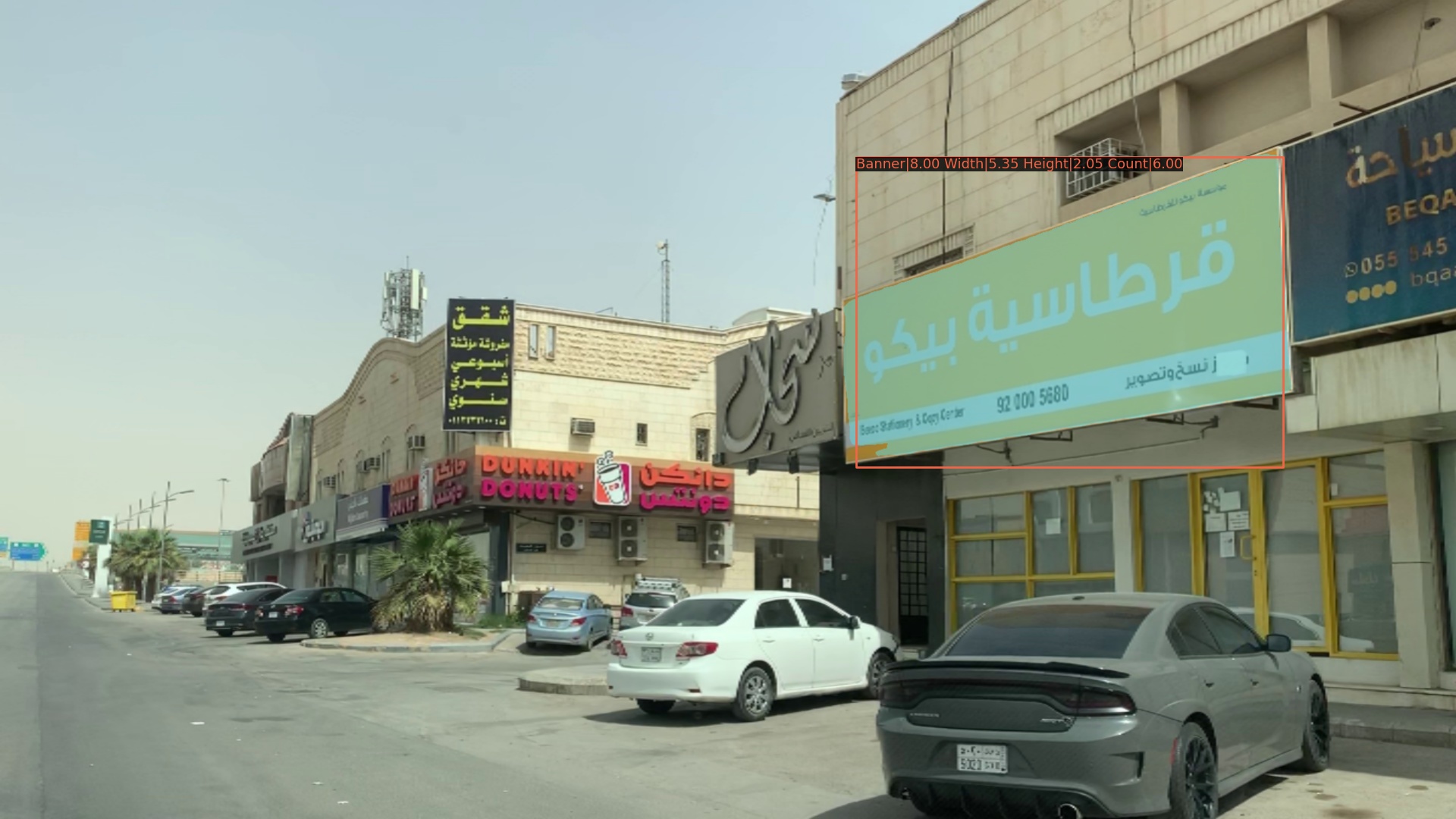}
\includegraphics[width=0.33\linewidth]{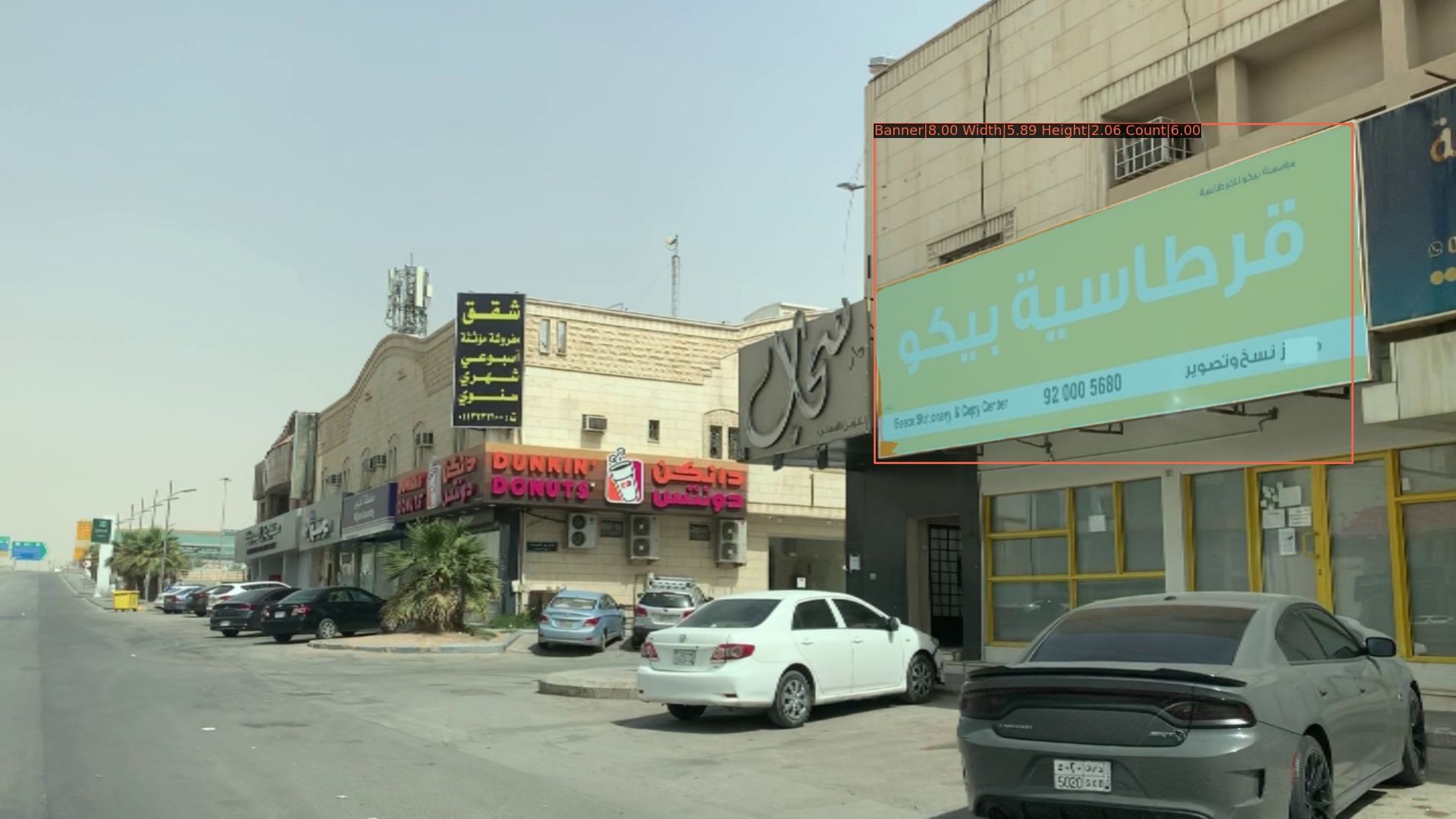}
\caption{The images present the semantic consistency within the video setting for model performance with object localization, segmentation, and height and width estimation.\\
I row    : Image1 Width: 5.54m Height: 1.84m Image2 Width: 5.54m Height: 1.84m Image3 Width: 5.48m Height: 1.62m \\II row : Image1 Width: 6.20m Height: 2.06m Image2 Width: 6.09m Height: 2.07m Image3 Width: 5.93 Height: 2.05m \\III row: Image1 Width: 5.88m Height: 2.11m Image2 Width: 6.22m Height: 2.02m Image3 Width: 6.13m Height: 1.97m \\IV row: Image1 Width: 5.53m Height: 2.09m Image2 Width: 5.35m Height: 2.05m Image3 Width: 5.89m Height: 2.06m}
\end{figure}

\section{Conclusion}
Our work shows the application of Deep Neural Networks using the segmentation of pixels into prototype masks to measure the dimension of a common object. We estimated the 2D dimensions of a quadrilateral object based on the pixel-level segmentation objects adapted to instances. Our framework is generic to enclosed polygons. During the experiments, we noticed that the system  would learn three cameras' intrinsic and extrinsic properties. We also validated the robustness of the PMODE model with semantic consistency in videos by applying the model in real-world scenarios.

In the future, we intend to perform keypoint detection-based object dimension estimation wherein the network implicitly learns the object's Depth without performing pixel-wise segmentation. Key points of the shop signage or an object would be the inputs to the neural network where it does not have to be concerned about the pixel details. 

We would like to express our sincere thanks to Muhammad AlQurishi for the valuable review and feedback on this paper.
\bibliographystyle{unsrt}
\bibliography{ref}

\end{document}